\pdfoutput=1
\documentclass[11pt,a4paper,logo,onecolumn]{accelerate}
\usepackage[authoryear,round]{natbib}
\usepackage{amsmath,amssymb,graphicx,booktabs,tabularx,array,float,xcolor,longtable}
\usepackage{hyperref,url,pifont}
\usepackage{microtype}
\usepackage{flafter,placeins,wrapfig,needspace}
\usepackage{ragged2e}
\usepackage{enumitem}
\usepackage[most]{tcolorbox}
\definecolor{tasknamefill}{HTML}{EFF6FC}
\definecolor{tasknameframe}{HTML}{BDD4E8}
\newtcbox{\tasknamebox}{on line,boxsep=0pt,left=1.3pt,right=1.3pt,top=0.35pt,bottom=0.35pt,arc=0.6pt,boxrule=0.3pt,colback=tasknamefill,colframe=tasknameframe,colupper=black,fontupper=\rmfamily\upshape}
\DeclareRobustCommand{\taskname}[1]{\tasknamebox{\vphantom{Ag}#1}}
\definecolor{insightteal}{HTML}{109D9F}
\definecolor{insightink}{HTML}{17334D}
\newtcolorbox{insightbox}[1]{enhanced,colback=insightteal!6!white,colframe=insightteal,boxrule=0pt,borderline west={1.4pt}{0pt}{insightteal},sharp corners,left=8pt,right=8pt,top=4pt,bottom=4pt,before skip=5pt,after skip=5pt,fonttitle=\bfseries,coltitle=insightink,colbacktitle=insightteal!6!white,title={#1},titlerule=0pt}
\newcommand{\observation}[1]{\ding{\numexpr171+#1\relax}}
\hypersetup{hidelinks,pdftitle={Accelerate Bench: Evaluating Agents for Application-Level Program Optimization}}
\definecolor{draftblue}{HTML}{3977B8}
\definecolor{draftorange}{HTML}{B86242}

\newcolumntype{Y}{>{\raggedright\arraybackslash}X}
\definecolor{revblue}{HTML}{000000}
\newcommand{\rev}[1]{{\color{revblue}#1}}
\newcommand{\VNsys}{16}

\newcommand{\VOutcomes}{512}
\newcommand{\VCorrect}{472}

\newcommand{\VCorrMin}{26}
\newcommand{\VCorrMax}{32}
\newcommand{\VBest}{67.4}
\newcommand{\VWorst}{25.5}
\newcommand{\VRefDropMin}{0.5}
\newcommand{\VRefDropMax}{2.4}

\newcommand{\VAtLeast}{346}
\newcommand{\VAtwoMin}{8}
\newcommand{\VAtwoMax}{23}
\newcommand{\VLeadAtwoMin}{22}
\newcommand{\VRestWord}{fourteen}
\newcommand{\VRestCorrect}{412}
\newcommand{\VRestBelow}{125}
\newcommand{\VRestAtwoMin}{8}
\newcommand{\VRestAtwoMax}{17}
\newcommand{\VOpusScore}{59.4}
\newcommand{\VOpusOwn}{65.6}
\newcommand{\VOpusLead}{10.2}
\newcommand{\VSonnetLeadLo}{13.4}
\newcommand{\VSonOpus}{8.0}
\newcommand{\VSonOpusLo}{0.0}
\newcommand{\VSonOpusHi}{17.3}
\newcommand{\VOpusSolLo}{0.1}
\newcommand{\VOpusSolHi}{18.2}
\newcommand{\VPairsResolved}{64}
\newcommand{\VPairsTotal}{120}
\newcommand{\VSubN}{29}
\newcommand{\VSubTau}{0.95}
\newcommand{\VSubLead}{13.7}
\newcommand{\VSubTauH}{0.97}
\newcommand{\VSubLeadH}{13.9}
\newcommand{\VStopNote}{The provider's safety classifiers stopped the Claude Opus 5.5 runs on \taskname{Graph sparsification}, \taskname{PRESTO} and \taskname{VPIC}, flagging their sessions as cybersecurity or biology content \citep{anthropic_refusals}. For these two categories, Claude Code re-runs a flagged request on Claude Opus 4.8 (cybersecurity) or Claude Opus 5 (biology) and continues the session on that model \citep{claudecode_fallback}; because our gateway sends every request to the evaluated model, this fallback could not take effect, and the three runs ended. Table~\ref{tab:system_main} counts them as failures with zero reward, so Claude Opus 5.5 scores 59.4\%; on its other 29 tasks, it scores 65.6\%. Evaluations differ here. Vals AI runs Claude Opus 5 with Claude Opus 4.8 as a fallback for refusals and reports Terminal-Bench 2.1 both with the fallback's results and with those tasks counted as failures \citep{vals2026opus5}; the Artificial Analysis Coding Agent Index scores an attempt completed by a fallback model like any other \citep{aa2026codingagent}, and \citet{davidovich2026permission} keep the vendor's fallback because it is the route an organization deploys. Taking the run of Claude Sonnet 5.5, the other Claude model in Table~\ref{tab:system_main}, on the same three tasks as such a fallback would give 65.2\%.}

\title{Accelerate Bench: Evaluating Agents for\\Application-Level Program Optimization}
\input{metadata}
\usepackage{catchfile}

\definecolor{gblue9}{RGB}{23,78,166}
\fancypagestyle{hyfirst}[firststyle]{\fancyfoot[L]{}\fancyfoot[R]{\footerfont\thepage}}
\hypersetup{pdfauthor={Bowen Liu, Yuhan Li, Nuo Chen, Weile Luo, Miao Peng, Qintong Zhang, Qifan Zhang, Jia Li, Ziqi Gao, Lydell Huang}}

\begin{document}
\raggedbottom
\begin{abstract}
Application optimization requires coding agents to discover useful changes and coordinate them across a complete program while preserving the required computation. We introduce \textbf{Accelerate Bench}, a benchmark of 32 CPU optimization tasks adapted from scientific applications and supercomputing competitions. Each task provides a working program and specifies the workload, correctness requirements, timing boundaries, and permitted edits, leaving agents to choose where and how to optimize. Public testing supports iterative development, and a separate verifier scores the final submission.
\rev{On benchmark version 1.1, we evaluate \VNsys{} coding agents, each pairing a model with a harness. Their Score@1 ranges from \VWorst\% to \VBest\%. Most of their programs are correct, and the size of the speedup separates them: the two leading agents are at least $2\times$ faster on \VLeadAtwoMin{} or more of the 32 tasks, while the other \VRestWord{} reach $2\times$ on \VRestAtwoMin{} to \VRestAtwoMax{} and leave \VRestBelow{} of their \VRestCorrect{} correct outcomes below $1.25\times$.} Controlled replays show why coordination across an application matters: identical source code runs between $0.29\times$ and $2.12\times$ depending only on build and runtime settings, and one code change is worth $1.01\times$ or $6.4\times$ depending on another. Accelerate Bench pairs executable workloads with agents' programs and trajectories to evaluate both their gains and the process behind them.
\end{abstract}

\maketitle\thispagestyle{hyfirst}
\suppressfloats[t]

\begin{figure}[t]
\centering\includegraphics[width=\linewidth]{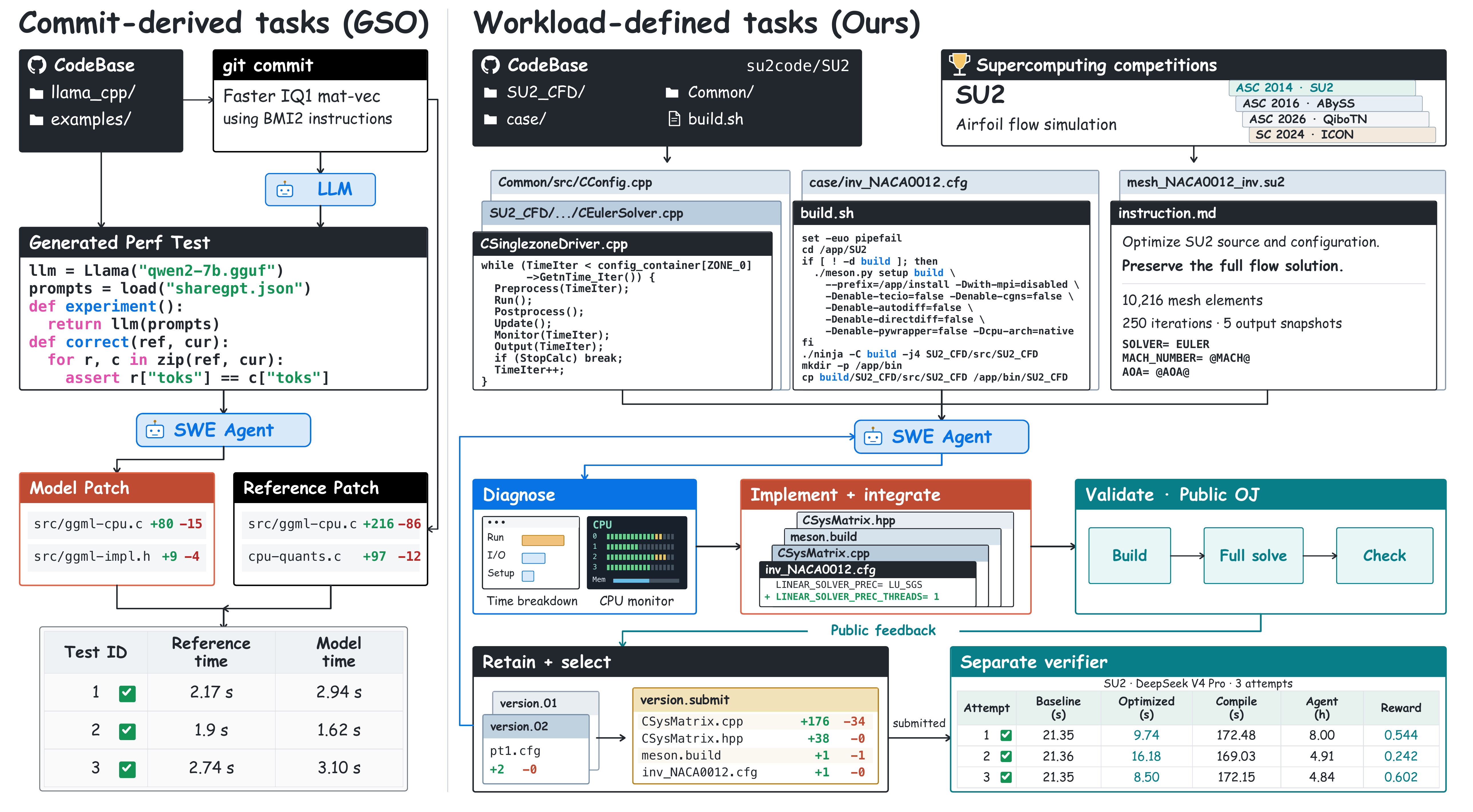}
\caption{\textbf{Application performance engineering in Accelerate Bench.} Left: GSO derives optimization tasks and reference patches from human commits \citep{gso}. Right: Accelerate Bench defines tasks by an application and its workload, illustrated with \taskname{SU2}. The agent diagnoses performance, modifies implementation, compiler options, and runtime configuration, and uses complete public tests to guide revisions and candidate selection. A separate verifier rebuilds and scores the final submission. GSO's table shows three example tests; the \taskname{SU2} table reports three DeepSeek V4 Pro runs. \taskname{SU2} runtime sums the mean solver times over three verification cases. Compile time is reported separately and excluded from reward; Agent time covers the agent-execution phase. The diagnostic displays illustrate the tools used in this workflow.}
\label{fig:overview}
\end{figure}

\section{Introduction}
Software performance engineering improves application performance by identifying and addressing inefficiencies in software and the way it executes~\citep{woodside2007future,leiserson2020plenty}. Effective optimization often requires interventions across multiple layers of the software stack, from compiler and build configurations to algorithms, data structures, dependencies, and hardware parallelism~\citep{leiserson2020plenty,balaprakash2018autotuning}. For an existing application, this is typically an iterative engineering process: measuring execution, locating performance bottlenecks, implementing targeted changes, and evaluating their end-to-end impact while preserving correctness~\citep{woodside2007future,balaprakash2018autotuning}.

Application-level optimization therefore extends beyond localized code editing. A valid patch may leave end-to-end runtime unchanged, and effective speedups can depend on interactions among changes made across different layers of the software stack. Success requires discovering where performance is lost, deciding where to intervene, and verifying that the resulting program is both correct and faster at the workload level.

Coding agents offer a path toward automating this process. Existing benchmarks such as SWE-Perf, GSO, and SWE-fficiency evaluate performance optimization in real software repositories using tasks derived from human optimization commits or pull requests~\citep{sweperf,gso,swefficiency}. These benchmarks anchor evaluation around known optimization instances, providing realistic repositories and expert optimization references.

We study a complementary setting: can an agent optimize an application when the required behavior and workload are specified, but no known optimization patch determines where or how the agent should intervene? Inspired by behavioral specification in ProgramBench~\citep{programbench}, we define optimization tasks by the computation that must be preserved and the workload that must become faster. Within a permitted edit scope, the agent may modify algorithms, implementation, build settings, dependencies, and runtime configuration.

We introduce \textbf{Accelerate Bench}, a benchmark of 32 CPU optimization tasks adapted from scientific applications and supercomputing competition problems. Each task provides a working initial program together with correctness requirements, timing boundaries, permitted edits, and a public testing interface. Agents iteratively edit, build, execute, and evaluate candidate programs, while a separate verifier scores the final submission. This setup requires agents to discover optimization opportunities and convert them into verified end-to-end acceleration.

\rev{We evaluate \VNsys{} coding agents on version 1.1 of Accelerate Bench, comparing correctness and end-to-end acceleration across the 32 tasks. Most of their programs are correct, so correctness alone does not separate the agents; the size of the speedup does. The two leading agents are at least $2\times$ faster on \VLeadAtwoMin{} or more tasks, whereas the other \VRestWord{} reach $2\times$ on \VRestAtwoMin{} to \VRestAtwoMax{} tasks and leave \VRestBelow{} of their \VRestCorrect{} correct outcomes below $1.25\times$.}

\rev{Accelerate Bench also records how agents optimize. From annotated trajectories of five agents on benchmark v1.0 and controlled replays of their programs, we ask what changes agents make to an application, how they test and develop candidate programs, and how optimization choices interact; the analysis characterizes optimization on these tasks rather than explaining the v1.1 ranking.} Our contributions are:
\begin{itemize}[leftmargin=*,itemsep=1pt,topsep=2pt]
\item \textbf{A workload-defined optimization benchmark.} Each of the 32 tasks fixes the computation, complete outputs, stateful checks such as checkpoint recovery, and timing boundaries; public tests support iteration, and a private verifier scores the final submission. In 14 planned controls on two tasks, every valid variant passes and all eight invalid variants fail the check they violate (Section~\ref{sec:benchmark}).
\item \rev{\textbf{An evaluation of \VNsys{} coding agents on benchmark v1.1.} Correctness alone does not separate the agents: each is correct on \VCorrMin{} to \VCorrMax{} of the 32 tasks, while Score@1 ranges from \VWorst\% to \VBest\% and the number of tasks with at least a $2\times$ speedup from \VAtwoMin{} to \VAtwoMax{} (Section~\ref{sec:results}).}
\item \textbf{An analysis of how agents optimize.} From 480 runs of five agents \rev{on benchmark v1.0} and controlled replays of their programs, we find that these agents mostly change the computation and rarely the build, although build and runtime settings alone move one unchanged \taskname{GROMACS} program between $0.29\times$ and $2.12\times$. Code changes also interact: one change in a \taskname{QiboTN} program is worth $1.01\times$ on its own and $6.4\times$ once another change is in place (Section~\ref{sec:analysis}).
\end{itemize}
We will release the versioned task packages, including public tests and verifiers.

\section{Benchmark Construction}\label{sec:benchmark}
Accelerate Bench comprises 32 CPU optimization tasks adapted from mature scientific applications across six established supercomputing competitions. We adapt each task into an executable workload running in a resource-constrained environment, with explicit permitted modifications, timing boundaries, and output requirements. We pair interactive public feedback during development with held-out private verification. Experienced finalists of the ASC Student Supercomputer Challenge reviewed the task designs. Appendix~\ref{sec:taskorigins} documents task origins and adaptations.

\subsection{Task Definition and Optimization Scope}
Unlike benchmarks built around isolated historical commits or patches~\citep{gso,sweperf}, we define each task by a complete application workload and its required outputs. Each task fixes the target computation and leaves agents to decide where and how to optimize. We evaluate performance across the full workflow, so local improvements count only if they reduce total runtime.

\begin{figure}[t]
\centering
\includegraphics[width=5.3in]{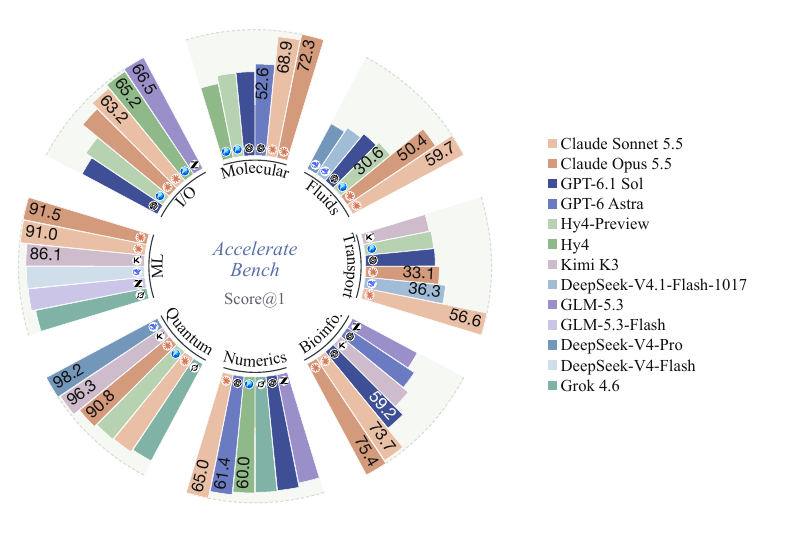}
\caption{\textbf{Score@1 by application area.} \rev{Benchmark v1.1, first run of the \VNsys{} agents in Table~\ref{tab:system_main};} top six agents per area, increasing clockwise. Labels give the top three scores (\%); each sector is scaled to its own maximum (dashed arc). \rev{Claude Opus 5.5's three runs stopped by the provider's safety system, on two Transport tasks and one Numerics task, count as zero (Table~\ref{tab:system_main}).}}
\label{fig:dataset_diversity}
\end{figure}
Each task provides a working baseline implementation, build and execution scripts, task instructions, and a public testing interface in a container with fixed CPU core and memory limits (Appendix~\ref{sec:inventory}). Agents may modify application source code, compiler options, and dependencies within the permitted directory, while preserving the required computations and outputs. Application-level optimization means reducing the end-to-end runtime of this prescribed workflow under these resource limits.

Figure~\ref{fig:overview} illustrates this workflow with \taskname{SU2}: agents optimize an airfoil flow simulation by editing solver routines, common libraries, and compilation flags. To pass, each run must complete all 250 iterations on the supplied 10,216-element mesh and produce five complete flow-field snapshots. The public test rebuilds the candidate, runs the simulation, and checks field residuals and run-to-run consistency against reference solutions. The verifier alone controls the validation meshes, physical configurations, and the checking code used to score submissions.

Timing boundaries follow each task's computational workflow, excluding one-time setup and build steps\rev{; end-to-end runtime in this paper means this timed workflow, and build time is reported separately}. For example, \taskname{GROMACS} times the \texttt{mdrun} simulation, excluding build time and topology preparation; the \taskname{HPCG-derived} task times sparse matrix assembly, initialization, multigrid solves, and output; and \taskname{paged generation} times model loading, autoregressive decoding, checkpoint saving, and recovery in a new process. Because these stages are timed and checked together, a faster kernel counts only in proportion to its share of the workflow, and a shortcut that skips a required output or recovery step fails correctness rather than earning a speedup. Appendix Table~\ref{tab:contracts} gives representative contracts, and Table~\ref{tab:resources} lists each task's CPU cores, memory, and time limits; every task package specifies its own outputs and checks.
\par

\subsection{Task Adaptation, Domain Coverage, and Workload Scaling}
The 32 tasks span eight application areas (Figure~\ref{fig:dataset_diversity}) curated from six supercomputing competitions: ASC, PAC, the ACM-China International Parallel Computing Challenge (IPCC), SCC at SC and ISC, and IndySCC (Appendix Table~\ref{tab:competition_sources}). Each task adapts a competition's target application and optimization challenge into a bounded, self-contained workload rather than reproducing the original input (Appendix~\ref{sec:taskorigins}). \rev{Twenty-nine tasks build on the competition's application, from full source trees to selected solver subtrees; three (\taskname{Paged LM generation}, \taskname{Tucker reconstruction} and \taskname{Graph sparsification}) are workflows we wrote around a competition's optimization theme.} We standardize dependencies, build scripts, and execution entry points in isolated containers, using inputs from canonical benchmarks, public datasets, or synthetic generators.

\paragraph{Workload scaling.}
Competition workloads typically run on multi-node clusters for hours. To allow agents to iteratively edit, build, and test within a practical evaluation budget, we adapt these workloads to single-node CPU environments by reducing input sizes and iteration counts. The adapted tasks retain the main computational stages, inter-stage dependencies, and numerical correctness criteria of the original applications (adaptation parameters in Appendix~\ref{sec:taskorigins}). \rev{Because private inputs are often larger than the public ones (Section~\ref{sec:grading}), a change that helps only on small inputs earns less. Timing covers the whole workflow, so savings in fixed costs such as process startup and library imports also count.}

\subsection{Public Tests, Private Verification, and Checks}\label{sec:grading}
Agents develop against a public test; a private verifier scores the submission (Figure~\ref{fig:overview}). During the run, the agent can call the public test repeatedly; it rebuilds the current program, runs it on a few public inputs (typically two), checks the outputs, and reports correctness, runtime and, in most tasks, the baseline runtime or speedup, or the failing step. The agent sees whether its program builds, produces the required outputs, and runs faster, but not the checking code. After the run, the hidden verifier rebuilds the submitted files and runs them on private inputs of the same kind, typically four. Most tasks include the public inputs and held-out inputs that are often larger or freshly randomized per evaluation, such as other Mach numbers and angles of attack for \taskname{SU2}. Only the verifier's repeated timings and checks set the reward (Section~\ref{sec:protocol}), so the score depends on inputs the agent has not seen.

To prevent invalid optimizations—such as skipping required computations or omitting output fields—the verifier enforces correctness checks. Submissions must meet task-specific numerical tolerances, generate complete output files, and satisfy stateful requirements such as recovering correctly from a saved checkpoint in a new process; only submissions passing all checks are scored on runtime. In 14 planned controls on \taskname{HPCG-derived} and \taskname{paged generation} (Appendix~\ref{sec:validation}), every valid variant passes and all eight invalid variants fail the check they violate: omitting the second solution fails the output-size check, and four attention variants that pass component tests fail application checks.

\section{Evaluation Metrics}\label{sec:protocol}
Using the task-provided implementation as the baseline, we measure the runtime agents save while preserving the required computation. All metrics weight the 32 tasks equally.

\paragraph{Runtime savings.}
For task $t$, let $B_t$ be the baseline runtime and $T_t(a)$ the runtime of a submitted program $a$, measured under the same timing rule. Let $C_t(a)\in\{0,1\}$ indicate whether the program passes the final correctness checks. For outcomes with a valid runtime, the reward is defined as
\begin{equation}
 r_t(a)=C_t(a)\max\left(0,1-\frac{T_t(a)}{B_t}\right).
 \label{eq:runtime_savings}
\end{equation}
Failed outcomes and those without a valid runtime receive zero. Thus, a correct 2$\times$ speedup earns 0.5 and a correct but no faster program earns zero; the reward is the fraction of baseline runtime saved.

\paragraph{Scored run.}
Each run gives the agent eight hours to inspect, edit, build and test the program before its final submission is scored. We denote the run of agent $m$ on task $t$ by $a^*_{m,t}$; all metrics and usage counters refer to its final submission. Our primary metric, \textbf{Score@1}, is the mean reward over tasks:
\begin{equation}
 \mathrm{Score@1}_m=\frac{1}{|\mathcal{T}|}\sum_{t\in\mathcal{T}}r_t(a^*_{m,t}).
 \label{eq:score_at_one}
\end{equation}

\rev{\paragraph{Scored runs in benchmark v1.1.}
Every agent runs once per task, and this run is scored; final programs are scored on the same machine (Appendix~\ref{sec:v11_changes}). GPT-6 Astra and Hy4-Preview run two more times to measure run-to-run variation (Appendix~\ref{sec:v11_variance}). A run that an infrastructure or model-API failure stops without a valid result receives no score and is replaced by a new run fixed in advance; failures caused by the agent, including incorrect output, out-of-memory termination and timeout, are scored. Three Claude Opus 5.5 runs that the provider's safety system stopped count as failures (Appendix~\ref{sec:v11_changes}). Appendix~\ref{sec:v10_results} gives the run selection of benchmark v1.0.}

\paragraph{Speedup targets.}
We report $A_\tau$, the fraction of tasks whose run is correct and achieves at least $\tau$-fold speedup. We use $\tau=1.25$ and $2$, corresponding to runtime savings of at least 20\% and 50\%, respectively\rev{; Appendix~\ref{sec:v11_thresholds} adds $\tau=5$ and $10$}. KernelBench uses a similar joint correctness-and-speed criterion \citep{kernelbench}.

\paragraph{Geometric mean speedup.}
\textbf{GM} is the geometric mean of the runs' speedups across tasks; incorrect or unaccelerated outcomes are assigned $1\times$. Replays of unchanged source on three tasks measure within 2\% of $1\times$, far below the $1.25\times$ target; Appendix~\ref{sec:diagnostic_metrics} reports this calibration together with sensitivity analyses for the scoring design and baseline choice. \rev{In v1.1, the unmodified programs of all 32 tasks, each judged once by the private verifier, measure between $0.923\times$ and $1.017\times$ (median $0.999\times$; Appendix~\ref{sec:config_baseline}).}

\section{Main Results}\label{sec:results}
\paragraph{Experimental setup.}
\rev{We compare \VNsys{} coding agents on the 32 tasks of benchmark v1.1 (Appendix~\ref{sec:v11_changes}). Each agent pairs a model with a harness: GPT-6 Astra, GPT-6.1 Sol and GPT-6.1 Terra use Codex, and the other models use Claude Code, so scores compare these combinations rather than base models (Section~\ref{sec:limitations}). Table~\ref{tab:system_main} reports performance; final programs are scored on the same machine. Benchmark v1.0 results are in Appendix~\ref{sec:v10_results}.}

\begin{table}[!t]
\centering\small\setlength{\tabcolsep}{2.5pt}\color{revblue}
\caption{\textbf{Performance on the 32 tasks of benchmark v1.1.} Every agent runs once per task, and final programs are scored on the same machine (Appendix~\ref{sec:v11_changes}); single runs do not order agents a few points apart (Appendix~\ref{sec:v11_resolution}). ``Config.\ ref.''\ recomputes Score@1 with the fixed configuration-only baseline (Appendix~\ref{sec:config_baseline}) as the reference. Scores and rates are percentages; GM is in $\times$. The v1.1 runs of Gemini-3.8-Flash are not yet complete, and it is not listed. $^\dagger$Runs on \taskname{Graph sparsification}, \taskname{PRESTO} and \taskname{VPIC}, stopped by the provider's safety classifiers (cybersecurity or biology flags), count as failures; on its other 29 tasks, Claude Opus 5.5 scores 65.6\% (Appendix~\ref{sec:v11_changes}). Bold marks maxima.}
\label{tab:system_main}
\begin{tabular*}{\linewidth}{@{\extracolsep{\fill}}llrrrrr@{}}
\toprule
Agent & Harness & Score@1 & Config.\ ref. & GM & $A_{1.25}$ & $A_2$\\
\midrule
\raisebox{-0.15ex}{\includegraphics[width=1em,height=1em,keepaspectratio]{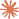}}\hspace{0.35em}Claude Sonnet 5.5 & Claude Code & \textbf{67.4} & \textbf{66.3} & 3.83 & \textbf{100.0} & \textbf{71.9} \\
\raisebox{-0.15ex}{\includegraphics[width=1em,height=1em,keepaspectratio]{figures/model-logos/claude-color.pdf}}\hspace{0.35em}Claude Opus 5.5$^\dagger$ & Claude Code & 59.4 & 58.3 & \textbf{3.94} & 84.4 & 68.8 \\
\raisebox{-0.15ex}{\includegraphics[width=1em,height=1em,keepaspectratio]{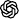}}\hspace{0.35em}GPT-6.1 Sol & Codex & 49.2 & 46.9 & 2.42 & 78.1 & 50.0 \\
\raisebox{-0.15ex}{\includegraphics[width=1em,height=1em,keepaspectratio]{figures/model-logos/openai.pdf}}\hspace{0.35em}GPT-6 Astra & Codex & 47.2 & 45.0 & 2.40 & 75.0 & 53.1 \\
\raisebox{-0.15ex}{\includegraphics[width=1em,height=1em,keepaspectratio]{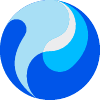}}\hspace{0.35em}Hy4-Preview & Claude Code & 45.2 & 42.9 & 2.31 & 71.9 & 50.0 \\
\raisebox{-0.15ex}{\includegraphics[width=1em,height=1em,keepaspectratio]{figures/model-logos/hunyuan-color.pdf}}\hspace{0.35em}Hy4 & Claude Code & 45.0 & 42.9 & 2.41 & 75.0 & 46.9 \\
\raisebox{-0.15ex}{\includegraphics[width=1em,height=1em,keepaspectratio]{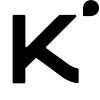}}\hspace{0.35em}Kimi K3 & Claude Code & 42.5 & 42.0 & 2.47 & 68.8 & 40.6 \\
\raisebox{-0.15ex}{\includegraphics[width=1em,height=1em,keepaspectratio]{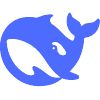}}\hspace{0.35em}DeepSeek-V4.1-Flash-1017 & Claude Code & 40.1 & 37.7 & 2.28 & 65.6 & 34.4 \\
\raisebox{-0.15ex}{\includegraphics[width=1em,height=1em,keepaspectratio]{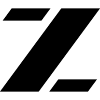}}\hspace{0.35em}GLM-5.3 & Claude Code & 39.7 & 38.3 & 2.20 & 65.6 & 40.6 \\
\raisebox{-0.15ex}{\includegraphics[width=1em,height=1em,keepaspectratio]{figures/model-logos/zai.pdf}}\hspace{0.35em}GLM-5.3-Flash & Claude Code & 39.3 & 37.0 & 2.20 & 68.8 & 34.4 \\
\raisebox{-0.15ex}{\includegraphics[width=1em,height=1em,keepaspectratio]{figures/model-logos/deepseek-color.pdf}}\hspace{0.35em}DeepSeek-V4-Pro & Claude Code & 37.9 & 36.1 & 2.19 & 62.5 & 34.4 \\
\raisebox{-0.15ex}{\includegraphics[width=1em,height=1em,keepaspectratio]{figures/model-logos/openai.pdf}}\hspace{0.35em}GPT-6.1 Terra & Codex & 37.9 & 36.7 & 1.92 & 75.0 & 25.0 \\
\raisebox{-0.15ex}{\includegraphics[width=1em,height=1em,keepaspectratio]{figures/model-logos/deepseek-color.pdf}}\hspace{0.35em}DeepSeek-V4-Flash & Claude Code & 34.1 & 31.8 & 1.92 & 56.2 & 34.4 \\
\raisebox{-0.15ex}{\includegraphics[width=1em,height=1em,keepaspectratio]{figures/model-logos/zai.pdf}}\hspace{0.35em}GLM-5.2 & Claude Code & 34.0 & 31.6 & 1.98 & 46.9 & 37.5 \\
\raisebox{-0.15ex}{\includegraphics[width=1em,height=1em,keepaspectratio]{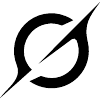}}\hspace{0.35em}Grok 4.6 & Claude Code & 33.9 & 31.7 & 2.13 & 50.0 & 34.4 \\
\raisebox{-0.15ex}{\includegraphics[width=1em,height=1em,keepaspectratio]{figures/model-logos/hunyuan-color.pdf}}\hspace{0.35em}Hy3 & Claude Code & 25.5 & 23.2 & 1.76 & 37.5 & 28.1 \\
\bottomrule
\end{tabular*}
\end{table}

\begin{figure}[!t]
\centering
\includegraphics[width=\linewidth]{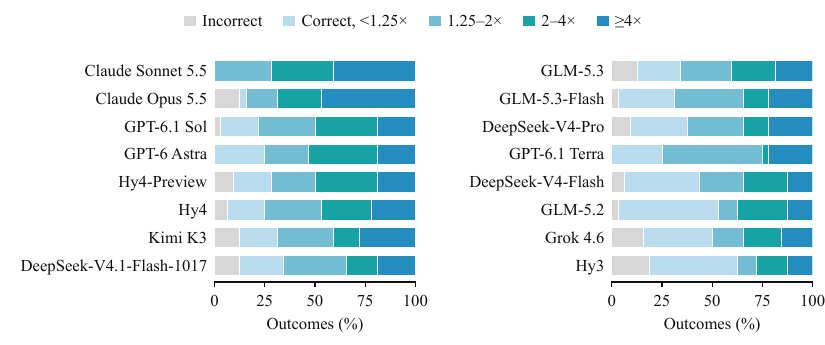}
\caption{\textbf{Correctness and acceleration.} Each bar splits an agent's 32 task outcomes into incorrect programs and correct programs binned by speedup over the baseline\rev{; benchmark v1.1, first run, agents ordered by Score@1. Claude Opus 5.5's three runs stopped by the provider's safety system count as incorrect (Table~\ref{tab:system_main})}.}
\label{fig:acceleration_distribution}
\end{figure}

\noindent\rev{\textbf{The two Claude agents lead; single runs do not order most other agents.} Claude Sonnet 5.5 and Claude Opus 5.5 reach a Score@1 of 67.4\% and \VOpusScore\%, at least \VOpusLead{} points above GPT-6.1 Sol (49.2\%). Three Claude Opus 5.5 runs that the provider's safety classifiers stopped for cybersecurity or biology content count as failures; on its other 29 tasks, it scores \VOpusOwn\% (Appendix~\ref{sec:v11_changes}). Paired task-bootstrap intervals put Claude Sonnet 5.5 at least \VSonnetLeadLo{} points ahead of every non-Claude agent and Claude Opus 5.5 \VOpusSolLo--\VOpusSolHi{} points ahead of GPT-6.1 Sol, but no two neighbors below them are separated. These intervals cover task variation only: re-scoring the same programs moves a single run's Score@1 by about 1.5 points, and independent runs of GPT-6 Astra and Hy4-Preview vary more (standard deviations of 2.5 and 4.4 points; Appendix~\ref{sec:v11_variance}), so one run does not order agents a few points apart. Without the three task-authored workflows or \taskname{HPCG-derived}, the order barely changes (Kendall's $\tau_b\geq\VSubTau$; Appendix~\ref{sec:v11_resolution}). With the fixed configuration-only baseline as the reference, each agent's Score@1 falls by \VRefDropMin--\VRefDropMax{} points (Appendix~\ref{sec:config_baseline}).}

\noindent\rev{\textbf{Most programs are correct; the size of the speedup separates agents.} Of the \VOutcomes{} agent--task outcomes, \VCorrect{} pass correctness checks, and \VAtLeast{} of these are also at least $1.25\times$ faster than the baseline (Figure~\ref{fig:acceleration_distribution}). Each agent is correct on \VCorrMin{} to \VCorrMax{} of the 32 tasks, so correctness alone does not order agents: GPT-6.1 Terra is correct on all 32 tasks but scores 37.9\%, below Kimi K3 at 42.5\% with 28. The two Claude agents are at least $2\times$ faster on \VLeadAtwoMin{} or more tasks. The other \VRestWord{} agents reach $2\times$ on \VRestAtwoMin{} to \VRestAtwoMax{} tasks, and \VRestBelow{} of their \VRestCorrect{} correct outcomes stay below $1.25\times$.}

\section{Analysis}\label{sec:analysis}
We answer the three questions from Section~1: what changes agents make to an application (Section~\ref{sec:rq1}), how they test and develop candidate programs (Section~\ref{sec:rq2}), and how optimization choices interact to affect end-to-end performance (Section~\ref{sec:mechanism_controls}).
\paragraph{Analysis sample.}\label{sec:analysis_data}
For this analysis, we run five agents, DeepSeek-V4.1-Flash, Grok 4.6, DeepSeek-V4-Pro, DeepSeek-V4-Flash, and Hy3, three times on each task at high reasoning effort, giving 480 runs. A trajectory is the complete record of a run's messages, tool calls, and outputs; a candidate is a program the agent sends to the public tests. GPT-5.6 Luna labels the optimization techniques in each run's final submission (476 runs), and DeepSeek-V4.1-Flash annotates every tool call's action, reason, and cited feedback, together with the relationships between changes (Appendix~\ref{sec:annotation_methods}). \rev{These runs use benchmark v1.0; the analysis characterizes how agents optimize on these tasks and does not explain differences among the v1.1 agents. Findings on how agents test and develop candidates are illustrative observations: each states the evidence it rests on, and none estimates how often a behavior occurs beyond the runs analyzed.}

\subsection{What Agents Change}\label{sec:rq1}
\begin{figure}[!htbp]
\setlength{\abovecaptionskip}{4pt}
\centering
\includegraphics[width=\linewidth]{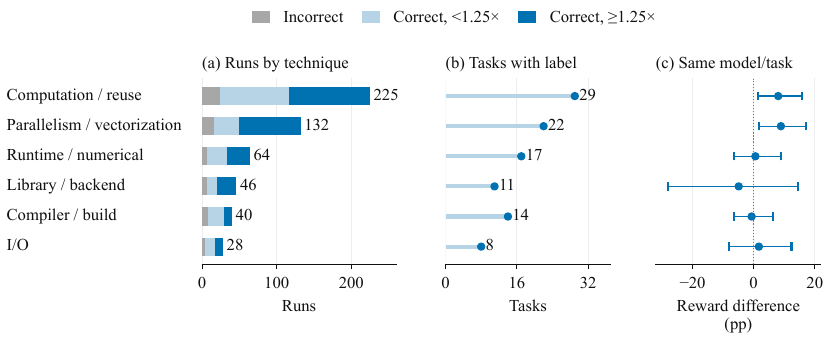}
\caption{\textbf{Optimization techniques and outcomes.} (a) Runs using each technique among 476 labeled runs, by final outcome. (b) Tasks in which each technique appears. (c) Mean reward difference between runs with and without the technique, from the same model on the same task; tasks are weighted equally, with 95\% task-bootstrap intervals.}
\label{fig:optimization_techniques}
\end{figure}

We group fifteen techniques found in final submissions into six categories (Figure~\ref{fig:optimization_techniques}).

\noindent\observation{1}\quad\textbf{Agents mostly change the computation, and rarely the build.} Computation or reuse changes appear in 225 of the 476 labeled runs, across 29 tasks, and parallelism or vectorization in 132 runs across 22 tasks. Compiler and build tuning appears in only 40 runs.

\noindent\observation{2}\quad\textbf{Runs with parallelism or vectorization most often reach the target speedup.} Of the runs using them, 62.9\% are correct and at least $1.25\times$ faster, the highest rate among the six categories, compared with 48.0\% for computation or reuse changes. Both keep correctness near 88\% (Table~\ref{tab:optimization_outcomes}).

\noindent\observation{3}\quad\textbf{Runs using either leading technique earn about eight more points of reward.} Against runs of the same model on the same task without them, mean reward is $8.0$ percentage points higher with computation or reuse changes (27 tasks; 95\% interval $[1.4,15.9]$) and $8.9$ points higher with parallelism or vectorization (19 tasks; $[1.9,17.0]$; Appendix~\ref{sec:technique_contrasts}).

\subsection{How Agents Test and Develop Candidates}\label{sec:rq2}
\begin{figure}[t]
\setlength{\abovecaptionskip}{4pt}
\centering
\includegraphics[width=\linewidth,trim=0 4bp 0 3bp,clip]{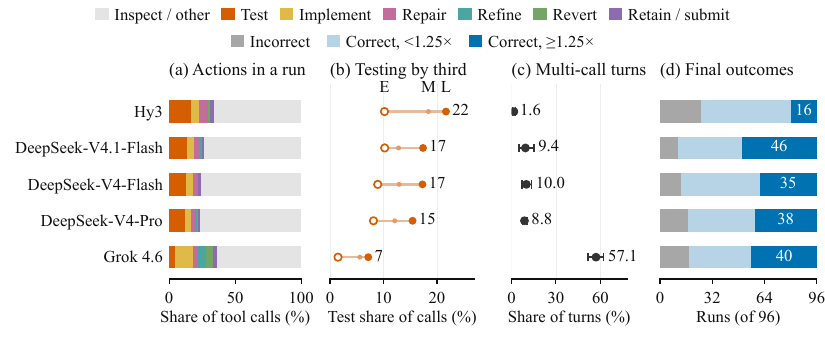}
\caption{\textbf{Every agent tests more late in a run, but Hy3, which tests most, reaches $1.25\times$ least often.} Each tool call has one decision label from the automatic annotator (Appendix~\ref{sec:annotation_methods}). (a) Mean share of a run's calls with each label. (b) Share of calls labeled \emph{test} in the early (E), middle (M), and late (L) third of a run; rows are sorted by L. (c) Share of tool-calling turns with two or more calls, with 95\% task-bootstrap intervals. (d) Outcomes of all 96 runs per agent, colored as in Figure~\ref{fig:optimization_techniques}. Definitions: Appendix~\ref{sec:trajectory_systems}; all trajectories: Figure~\ref{fig:trajectory_atlas}.}
\label{fig:trajectory_overview}
\end{figure}

\noindent\observation{1}\quad\textbf{All agents test more as their runs progress, but more testing does not guarantee success.} Every agent's share of test calls rises from the early to the late third of its trajectory (Figure~\ref{fig:trajectory_overview}b). Hy3 has the highest late-stage share, 21.7\%, yet reaches $1.25\times$ in only 16 of 96 runs, the fewest (Figure~\ref{fig:trajectory_overview}d).

\begin{figure}[t]
\centering
\includegraphics[width=\linewidth,trim=0 10bp 0 8bp,clip]{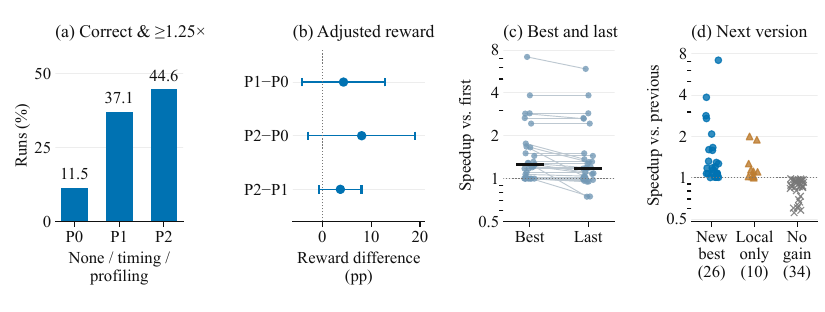}
\caption{\textbf{Timing and profiling, and search progress.} (a) Share of runs correct and at least $1.25\times$ faster, for runs with no timing or profiling command (P0), timing only (P1), or profiling (P2). (b) Reward differences adjusted for task and model, with 95\% task-cluster intervals. (c) Best and last correct candidates versus the first in 31 histories; lines join one history, black bars mark medians\rev{, and most candidates have a single timing}. (d) Speedup over the preceding correct candidate in 70 pairs with different source; ``Local only'' is faster without setting a new best. Speedup axes are logarithmic.}
\label{fig:workflow_summary}
\label{fig:optimization_diagnosis}
\label{fig:rq2_iteration}
\end{figure}

\noindent\observation{2}\quad\rev{\textbf{Most runs measure performance; much of their higher success reflects which tasks and models measure.}}\label{sec:workflow_measurement} Of the 480 runs, only 61 contain no detected timing or profiling command (P0), 251 contain timing commands without profiling (P1), and 168 contain profiling-related commands (P2). The fractions that are correct and at least $1.25\times$ faster are 11.5\%, 37.1\%, and 44.6\%, respectively. Much of this gap reflects which tasks and models measure: after adjusting for task and model, the P1--P0 reward difference is $+4.4$ percentage points (95\% interval $[-4.0,12.8]$; Figure~\ref{fig:workflow_summary}a--b, Appendix~\ref{sec:measurement_groups}).

\noindent\observation{3}\quad\textbf{Iteration pays off, but not monotonically.}\label{sec:workflow_iteration} We follow 31 version histories, runs with at least two correct candidates under one timing protocol, across seven applications. \rev{A run enters this set when it meets that condition and its candidate sources can be checked; no outcome measure is used. The counts below describe these 31 histories and are not estimates for all runs.} \rev{The last correct candidate is more than $1.05\times$ faster than the first in nineteen histories.} Progress is uneven: of 70 consecutive correct-candidate pairs with different source, 36 are faster than their predecessor, but only 26 set a new best (Figure~\ref{fig:workflow_summary}c--d; Appendix~\ref{sec:version_histories}). \rev{Agents also manage candidates explicitly: restore or select operations appear in 206 of 480 trajectories, and 93.0\% of the 583 turns with them cite earlier feedback (Appendix~\ref{sec:trajectory_decisions}).}

\subsection{How Optimization Choices Interact}\label{sec:mechanism_controls}
Controlled replays of submitted programs show that a change's gain depends on the settings and other changes around it, and sampled trajectories show \rev{how often agents meet such dependencies}.

\noindent\observation{1}\quad\textbf{The same source can run slower or twice as fast, depending only on build and runtime settings.} In one \taskname{GROMACS} run, the submission changes only the build and runtime configuration. Replaying this source under four settings, all correct, gives speedups from $0.289\times$ to $2.123\times$: AVX2 vectorization reaches $2.123\times$ with four threads but $0.709\times$ with one (Table~\ref{tab:gromacs_control}).

\noindent\observation{2}\quad\textbf{One code change can be worth $1.01\times$ or $6.4\times$, depending on another.} In a DeepSeek-V4-Flash \taskname{QiboTN} program, we switch three changes on and off in a full factorial design: broader dispatch to a fast exact simulator (B), reuse of a computed state (R), and deferred loading of tensor libraries (L). \rev{The replay simulates six quantum circuits of 9 to 14 qubits, each in a fresh process, and the timing includes process startup and library imports.} Deferred loading alone improves runtime by only $1.01\times$, but by 6.3--6.4$\times$ once B is on\rev{, because deferred loading avoids importing the tensor libraries only when the execution path does not need them. B and L together make the program 9.6--9.7$\times$ faster than with all three off; R adds nothing measurable on this workload} (three rounds; Appendix~\ref{sec:qibo_control}).

\noindent\observation{3}\quad\textbf{An optimization helps only if the application executes it.} In an \taskname{HPL} run, DeepSeek-V4-Flash optimizes \texttt{HPL\_dgemm.c}, then finds by comparing binaries that its executable is unchanged because the application calls \texttt{cblas\_dgemm}; it moves the optimization to that entry point (Appendix~\ref{sec:mechanism_evidence}).

\noindent\observation{4}\quad\rev{\textbf{Most runs meet dependencies across layers, and about half meet one that affects an optimization.}} We use the trajectory annotations to locate candidate evidence and review randomly sampled trajectories, both with and without annotated dependencies: a reviewer confirms a dependency in 34 of 40 flagged and 15 of 40 unflagged trajectories. Weighting the two strata by size, in an estimated 72\% of the 480 runs (95\% interval 63--80\%), tool output shows that a change's effect or correctness depended on build settings, runtime settings, or the code path actually executed\rev{. This estimate includes dependencies that only concern getting the program to build or run in the container; in 52\% (41--63\%), the dependency concerns an optimization's speed or correctness} (Appendix~\ref{sec:dependency_review}). In a \taskname{Palabos} run, the agent's OpenMP program runs for 11 minutes with the default thread count, which sees all 192 host CPUs, and for 1.5\,s with four threads. In \taskname{SIESTA}, a build with \texttt{-march=native} differs from the reference in 662 of 868 values until fused multiply--add contraction is disabled.

\section{Related Work}\label{sec:related}
\paragraph{Code optimization benchmarks.}
Earlier benchmarks measure program efficiency at function scale: PIE studies performance-improving edits to competitive-programming solutions, EffiBench, Mercury and ECCO compare the runtime of generated solutions under correctness tests, EvalPerf and ENAMEL add inputs that expose inefficient algorithms and compare against graded or expert references, and ParEval tests whether models write correct parallel code \citep{pie,effibench,mercury,ecco,evalperf,enamel,pareval}. SWE-Perf supplies target functions and measures test runtimes; GSO compares repository optimizations with expert implementations; SWE-fficiency asks agents to locate bottlenecks for a supplied workload \citep{sweperf,gso,swefficiency}. A cross-machine replay of their reference patches finds that many tasks do not consistently satisfy the original validity rules \citep{perf_reliability}; our timing calibration and scoring sensitivity analyses address the same concern (Appendix~\ref{sec:diagnostic_metrics}). KernelBench targets efficient GPU operators and model forward passes, and AlgoTune evaluates numerical solvers through a problem--solution interface; both score speedup over a reference implementation of a self-contained function \citep{kernelbench,algotune}. In production, ECO selects code through fleet-wide profiling and checks model-generated performance edits with tests and post-deployment monitoring \citep{eco}. Accelerate Bench instead fixes one complete application workflow for all agents and times it under application-specific correctness constraints; as in SWE-fficiency, agents must first find where runtime is spent, and they may change algorithms, implementations, and build and runtime settings.

\paragraph{Long-horizon agent evaluation.}
RE-Bench is closest in format: agents improve working starting solutions within budgets of up to eight hours, and several of its ML research-engineering tasks score runtime reductions against expert references \citep{rebench}. It normalizes scores between the starting and expert solutions; we report runtime ratios to the provided baseline and verify final submissions on private inputs. AlphaEvolve improves programs through model-proposed edits scored by automated evaluators \citep{alphaevolve}. ALE-Bench evaluates iterative algorithm engineering through public testing and final private evaluation \citep{alebench}. AutoLab studies sustained improvement of working implementations, and MLS-Bench tests whether improvements to ML components generalize across settings and scales \citep{autolab,mlsbench}. Accelerate Bench applies this iterative loop to application-level optimization and records the candidates, timings and feedback behind each result.

\paragraph{Scientific coding benchmarks.}
SciCode evaluates scientific reasoning and code synthesis, while ScienceAgentBench asks agents to implement data-driven tasks from research publications \citep{scicode,scienceagentbench}. AInsteinBench evaluates development in scientific repositories; SWE-bench Science covers scientific software repair and integration \citep{ainsteinbench,swebscience}. Accelerate Bench starts from working applications and asks agents to accelerate the prescribed computations, subject to constraints on numerical outputs and workflow behavior.

\section{Limitations}\label{sec:limitations}
\paragraph{Task coverage and baselines.}
The evaluation covers CPU workflows at the specified scales and resource budgets. Speedups are relative to task-provided baselines with varying degrees of tuning; they measure improvement from those starting points. \rev{Some speedups exploit structure in the specified workload, such as the \taskname{HPCG-derived} operator, which admits a direct solver; the tasks measure whether agents find and realize such opportunities, and} results at other scales, on other hardware, and in production environments require separate evaluation.

\paragraph{Agent and execution settings.}
As in other agent benchmarks, each result reflects a model running in its harness, so score differences combine model capability with harness behavior, inference compute, and execution conditions. Every agent receives the same task container, resource and time limits, and timing and correctness rules, and scores are runtime ratios to the task baseline. \rev{Table~\ref{tab:system_main} reports one run per agent, and repeated runs cover two agents (Appendix~\ref{sec:v11_variance}).} Separating models from harnesses and measuring sensitivity to hardware placement are natural extensions that the retained programs and containers support.

\rev{\paragraph{Benchmark versions.}
Results of v1.0 and v1.1 are reported separately and never combined; v1.1 revised the task statements and fixed several verifier, timing and environment problems (Appendix~\ref{sec:v11_changes}). The trajectory analyses use v1.0 runs.}

\paragraph{Trajectory samples.}
The trajectory analyses cover five agents. Source labels combine model-assisted annotation with targeted source checks, and prevalence estimates rest on reviewed samples. Version histories cover 31 runs in seven applications, most candidates have one timing observation, and controlled experiments establish effects for the tested artifacts, workloads, and settings.

\section{Conclusion}\label{sec:conclusion}
Accelerate Bench evaluates application-level optimization on 32 workload-defined CPU tasks with public feedback and private verification. \rev{On benchmark v1.1, most programs of the \VNsys{} agents are correct, and the size of the speedup separates the agents: the best reaches a Score@1 of \VBest\% in one run, and the two leading agents are at least $2\times$ faster on \VLeadAtwoMin{} or more tasks, the others on \VRestAtwoMin{} to \VRestAtwoMax{}.} \rev{In the analyzed v1.0 runs,} computation or reuse changes appear in 225 of 476 labeled runs, compiler and build tuning in only 40, yet controlled replays show that gains depend on build and runtime settings and on other code changes: identical source runs from $0.29\times$ to $2.12\times$, \rev{one code change is worth $1.01\times$ or $6.4\times$ depending on another, and in an estimated 52\% of analyzed runs, tool output shows such a dependency affecting an optimization's speed or correctness.}
\label{maintext:end}

\clearpage
\section*{Reproducibility Statement}
Evaluation metrics, run-inclusion and aggregation rules, task resources, and model configurations are described in Section~\ref{sec:protocol} and Appendices~\ref{sec:inventory}--\ref{sec:diagnostic_metrics}; Appendix~\ref{sec:evidence} lists the data snapshot behind each result. Each score is stored with its task and verifier versions, the scored program, and the execution settings; process analyses use retained trajectories and evaluator responses. Reproduction requires the same workload definitions, task and verifier versions, execution resources, and agent settings. We will release the versioned task packages, including public tests and verifiers, with the published version of the paper.

\section*{AI Use Statement}
\rev{The coding agents} evaluated in this paper are the objects of study. Beyond them, we used large language models in two roles.

\textbf{As research instruments.} GPT-5.6 Luna labels the optimization techniques in each run's final implementation, and DeepSeek-V4.1-Flash annotates every tool call in the 480 analyzed trajectories (Appendix~\ref{sec:annotation_methods}). These labels support the analyses in Section~\ref{sec:analysis}; prevalence estimates come from a separate review of random samples rather than from label counts (Appendix~\ref{sec:dependency_review}).

\textbf{As research and writing assistants.} AI assistants supported literature search, discussion of benchmark positioning and analysis design, drafting and editing of the manuscript, and code for data processing, figure generation, and evaluation infrastructure.

The authors reviewed all AI-assisted outputs, checked each citation against its source, and verified the reported numbers against the frozen evaluation outputs. The authors take full responsibility for the paper's content.

\bibliographystyle{iclr2027_conference}
\bibliography{paper_references}

\appendix
\begingroup\color{revblue}
\section{Benchmark v1.1}\label{sec:v11_changes}
After the v1.0 experiments, we checked every task statement against its verifier, and student supercomputing competition finalists reviewed the task statements again. Version 1.1 rewrites the statements of all 32 tasks and fixes several verification, timing and environment problems. The reward of Equation~\ref{eq:runtime_savings} and the task weights are unchanged. Task-by-task changes are listed in the v1.1 release notes. Each version's task packages, verifiers and results are released separately, and scores from different versions are never combined.

\paragraph{Task statements.}
Every statement now specifies the timing scope, repetition rules, reward formula, numerical tolerances and rejection conditions, together with the permitted optimization scope and the computation that must be preserved.

\paragraph{Task selection.}
Tasks start from the applications and optimization themes of six supercomputing competitions, chosen by their organizers as recognized optimization challenges. We keep a task only when it runs on one node with four to eight cores and at most 32\,GiB of memory, its outputs can be checked against scientific tolerances and required workflow steps such as checkpoint recovery, and the verifier can time it against its unmodified program. Coverage of application domains and bottleneck types guides the final choice. Three candidate tasks were removed for every agent because they could not be scored: \taskname{ABINIT} lacked a usable application-runtime evaluation, \taskname{block15} a usable performance baseline, and \taskname{SolverChallenge24} had an unresolved timing-resolution problem. Their verifiers gave every program zero reward by construction, so the removal does not depend on any agent's result.

\paragraph{Execution environment.}
Version 1.1 runs on a different evaluation platform from v1.0 (Appendix~\ref{sec:eval_protocol}). Private verification runs in a sandbox separate from the agent and from the public tests. The unmodified program and the candidate are timed alternately in the same private sandbox, with the order rules of v1.0; \taskname{HPCG-derived} now runs one warm-up and seven timings per side. For Table~\ref{tab:system_main}, final programs are scored once, with no model calls, on the same machine with two AMD EPYC 9K84 processors: the private sandbox is pinned to the four or eight physical cores of one NUMA node that the task allocates, without their SMT siblings, with the task's memory limit and no network access. Appendix~\ref{sec:v11_variance} measures the timing error.

\paragraph{Agents and run validity.}
All agents request high reasoning effort; in v1.0, Kimi K3 and GLM-5.3 requested \texttt{max}. GPT-6 Astra, GPT-6.1 Sol and GPT-6.1 Terra run in Codex, and all other models, including Grok 4.6 and Gemini-3.8-Flash, run in Claude Code (v1.0 used Grok Build and Gemini CLI for these two). A run stopped by an infrastructure failure (for example a lost cloud job or a failed image build) or by a model-API failure after the client's retries receives no score and is replaced by a new run fixed in advance; the replacement's result is used whatever it is. Failures caused by the agent are scored: incorrect output, out-of-memory termination, context or output beyond the model's limit, and timeout, where the verifier result produced after the timeout counts. Every run, including failed and replaced ones, is released with the benchmark.

\paragraph{Runs stopped by the provider's safety system.}
\VStopNote{}

\subsection{Speedup Thresholds}\label{sec:v11_thresholds}
Table~\ref{tab:v11_thresholds} extends $A_\tau$ to $\tau=5$ and $10$. Because the reward saves a fraction of runtime, $2\times$ earns 0.5, $10\times$ 0.9 and $100\times$ 0.99, so large speedups are compressed in Score@1; the thresholds show them directly.

\begin{table}[!ht]
\centering\small\color{revblue}
\caption{\textbf{Share of tasks with a correct program at least $\tau$ times faster ($A_\tau$, \%), benchmark v1.1, first run.} $\tau=1.25,2,5,10$ correspond to runtime savings of at least 20\%, 50\%, 80\% and 90\%.}
\label{tab:v11_thresholds}
\begin{tabular*}{\linewidth}{@{\extracolsep{\fill}}lrrrr@{}}
\toprule
Agent & $A_{1.25}$ & $A_2$ & $A_5$ & $A_{10}$\\
\midrule
Claude Sonnet 5.5 & 100.0 & 71.9 & 34.4 & 12.5 \\
Claude Opus 5.5 & 84.4 & 68.8 & 34.4 & 15.6 \\
GPT-6.1 Sol & 78.1 & 50.0 & 15.6 & 9.4 \\
GPT-6 Astra & 75.0 & 53.1 & 12.5 & 9.4 \\
Hy4-Preview & 71.9 & 50.0 & 18.8 & 6.2 \\
Hy4 & 75.0 & 46.9 & 15.6 & 6.2 \\
Kimi K3 & 68.8 & 40.6 & 21.9 & 12.5 \\
DeepSeek-V4.1-Flash-1017 & 65.6 & 34.4 & 18.8 & 9.4 \\
GLM-5.3 & 65.6 & 40.6 & 12.5 & 9.4 \\
GLM-5.3-Flash & 68.8 & 34.4 & 15.6 & 6.2 \\
DeepSeek-V4-Pro & 62.5 & 34.4 & 9.4 & 9.4 \\
GPT-6.1 Terra & 75.0 & 25.0 & 18.8 & 3.1 \\
DeepSeek-V4-Flash & 56.2 & 34.4 & 6.2 & 6.2 \\
GLM-5.2 & 46.9 & 37.5 & 12.5 & 9.4 \\
Grok 4.6 & 50.0 & 34.4 & 9.4 & 9.4 \\
Hy3 & 37.5 & 28.1 & 12.5 & 9.4 \\
\bottomrule
\end{tabular*}
\end{table}

\subsection{Configuration-Only Baseline}\label{sec:config_baseline}
This baseline measures what standard build and runtime configuration achieves without source changes. The recipe was fixed before any run. It changes only build and runtime settings, never source code, tolerances, solvers, block sizes or other numerical and algorithmic parameters, and adds three cumulative candidates: R1, the maximum optimization level with the native instruction set (\texttt{-O3 -march=native}), without options that change floating-point semantics such as \texttt{-ffast-math}; R2, the application's own SIMD or vectorization build option at the highest level the evaluation CPU supports and the task permits; R3, thread or process counts equal to the allocated cores with thread pinning, only where the task permits it and the submitted files control the timed run. The public tests select the fastest passing candidate among the unmodified program and the recipes; a simpler configuration within 2\% of the fastest is preferred. The selected configuration is judged once by the unmodified private verifier, with the same procedure as an agent's final submission and no model calls. \taskname{Graph sparsification}, \taskname{Paged LM generation} and \taskname{QiboTN} are pure Python programs whose verifiers do not compile submitted code and fix the thread count, so the recipe does not apply and their baseline is the unmodified program. Each unmodified program is also judged once as a control.

All 32 configuration results are correct, but only \taskname{GROMACS} exceeds $1.25\times$: changing its SIMD option from the default \texttt{None} to the highest permitted level, \texttt{AVX2\_256}, with \texttt{-march=native} gives $2.23\times$. The next highest are \taskname{ThunderSVM} at $1.249\times$ and \taskname{Palabos} at $1.16\times$. On 23 tasks the baseline is the unmodified program: the three pure Python tasks; \taskname{SU2}, whose verifier build already uses \texttt{-O3} with the native instruction set; \taskname{wtdbg2}, whose public tests report no runtime; and 18 tasks where no candidate both passed the public tests and ran more than 2\% faster than the unmodified program. On 8 of these 18, R1 changed the numerical output and failed the full output comparison. The unmodified controls measure $0.923\times$--$1.017\times$ (median $0.999\times$), a Score@1 of 0.40\%; the configuration baseline scores 3.85\%.

With this baseline as the reference on the nine tasks where it changes the configuration (floored at $1\times$), GPT-6 Astra's three-run Score@1 falls from 50.59\% to 48.47\% (runs: 50.87\%, 48.49\%, 46.04\%), and $A_{1.25}$ and $A_2$ each lose one task per run. Of the 2.12-point drop, 1.70 comes from \taskname{GROMACS}, where GPT-6 Astra's three runs reach $2.18\times$--$2.21\times$, close to the baseline's $2.23\times$. The largest speedups do not come from configuration: on \taskname{HPCG-derived}, \taskname{QiboTN} and \taskname{Paged LM generation}, the baseline is $0.995\times$, $1.009\times$ and $0.990\times$, while GPT-6 Astra's three runs reach $28.9\times$--$36.7\times$, $7.9\times$--$11.5\times$ and $11.1\times$--$11.9\times$. Table~\ref{tab:system_main} applies the same reference to the first run of every agent.

\begin{table}[!ht]
\centering\scriptsize\color{revblue}
\caption{\textbf{Fixed configuration-only baseline on benchmark v1.1.} ``Selected'' is the recipe chosen on the public tests: R1 maximum optimization level with native instruction set; R2 adds the application's own SIMD build option; R3 adds thread or process counts equal to the allocated cores with pinning; ``unmodified'' means no recipe beat the unmodified program by 2\% on the public tests; n/a marks pure-Python tasks without a configurable build or thread setting; $^\ast$\taskname{SU2}'s verifier build already uses \texttt{-O3} with the native instruction set, and \taskname{wtdbg2}'s public tests report no runtime. Speedups are single private-verifier measurements: the unmodified program (control), the selected configuration, and the reference used for the ``Config.\ ref.''\ column of Table~\ref{tab:system_main} (the configuration's speedup where a recipe is selected, floored at $1\times$).}
\label{tab:v11_config_baseline}
\begin{tabular*}{\linewidth}{@{\extracolsep{\fill}}llrrr@{}}
\toprule
Task & Selected & Control & Configuration & Reference\\
\midrule
\taskname{ABySS} & R1 & 0.923 & 0.991 & 1.000 \\
\taskname{DataLife} & unmodified & 0.943 & 1.035 & 1.000 \\
\taskname{Fluidity} & unmodified & 0.978 & 1.020 & 1.000 \\
\taskname{GPAW} & unmodified & 0.995 & 0.992 & 1.000 \\
\taskname{Graph sparsification} & n/a & 1.017 & 1.017 & 1.000 \\
\taskname{GROMACS} & R2 & 1.006 & 2.225 & 2.225 \\
\taskname{HPCG-derived} & unmodified & 1.001 & 0.995 & 1.000 \\
\taskname{HPL-derived} & R1 & 1.010 & 1.022 & 1.022 \\
\taskname{ICON} & unmodified & 1.005 & 0.994 & 1.000 \\
\taskname{IO500-derived} & R1 & 0.999 & 1.017 & 1.017 \\
\taskname{LAMMPS} & unmodified & 0.998 & 0.998 & 1.000 \\
\taskname{MHM2} & unmodified & 1.017 & 1.044 & 1.000 \\
\taskname{miniVite} & R1 & 1.007 & 0.992 & 1.000 \\
\taskname{MPAS-A} & unmodified & 0.993 & 1.006 & 1.000 \\
\taskname{MrBayes} & R1 & 0.991 & 1.055 & 1.055 \\
\taskname{NEMO} & unmodified & 0.991 & 0.975 & 1.000 \\
\taskname{NUFFT imaging} & unmodified & 1.006 & 1.030 & 1.000 \\
\taskname{OpenFOAM} & unmodified & 0.992 & 0.973 & 1.000 \\
\taskname{OpenMC} & unmodified & 1.011 & 1.004 & 1.000 \\
\taskname{Paged LM generation} & n/a & 0.990 & 0.990 & 1.000 \\
\taskname{Palabos} & R1 & 1.011 & 1.162 & 1.162 \\
\taskname{PRESTO} & unmodified & 0.986 & 1.016 & 1.000 \\
\taskname{QiboTN} & n/a & 1.009 & 1.009 & 1.000 \\
\taskname{Quantum ESPRESSO} & unmodified & 0.991 & 0.992 & 1.000 \\
\taskname{QuEST} & R1 & 0.940 & 0.997 & 1.000 \\
\taskname{ShengBTE} & unmodified & 1.014 & 1.008 & 1.000 \\
\taskname{SIESTA} & unmodified & 0.976 & 0.998 & 1.000 \\
\taskname{SU2} & unmodified$^\ast$ & 0.997 & 0.997 & 1.000 \\
\taskname{ThunderSVM} & R1 & 0.999 & 1.249 & 1.249 \\
\taskname{Tucker reconstruction} & unmodified & 0.984 & 0.971 & 1.000 \\
\taskname{VPIC} & unmodified & 1.014 & 1.070 & 1.000 \\
\taskname{wtdbg2} & unmodified$^\ast$ & 1.003 & 1.003 & 1.000 \\
\bottomrule
\end{tabular*}
\end{table}

\subsection{Run-to-Run Variance and Timing Noise}\label{sec:v11_variance}
To separate the variation from agent search from the variation from timing a fixed program, we froze the final submissions of GPT-6 Astra's three runs (96 submissions) and the 77 submissions of Hy4-Preview's three runs that had been verified at the time, and judged each twice more with the unmodified private verifier, with no model calls (Table~\ref{tab:v11_variance}).

\begin{table}[!ht]
\centering\small
\caption{\textbf{Variation from re-judging fixed submissions and between independent runs (benchmark v1.1).} Standard deviation of Score@1 between two additional judgments of every fixed submission; it includes both timing and the correctness changes in the last row, and the 95\% interval resamples tasks. Speedup variation uses submissions judged correct in both judgments.}
\label{tab:v11_variance}
\begin{tabular*}{\linewidth}{@{\extracolsep{\fill}}lrr@{}}
\toprule
 & GPT-6 Astra (96) & Hy4-Preview (77)\\
\midrule
SD of Score@1 when re-judged (points) & 1.45 (0.70--2.04) & 1.57\\
SD of Score@1 between three independent runs (points) & 2.48 & 4.42\\
Median coefficient of variation of speedup & 3.0\% & 3.2\%\\
Submissions whose correctness changed between judgments & 3/96 & 7/77\\
\bottomrule
\end{tabular*}
\end{table}

Independent runs vary more than re-judged submissions. For GPT-6 Astra, by point estimate, re-judging accounts for about one third of the variance between independent runs and the agent's search for the rest; with three runs, this split is imprecise (95\% interval of the ratio 0.07--3.30). Re-judging variation has two parts. Most programs time stably: the median coefficient of variation of speedup is about 3\%. In addition, the correctness of ten submissions changed between judgments: seven produced numerical outputs within the tolerance in one judgment and outside it in the other, and three failed to run once. The verifier's checks and tolerances were the same in both judgments; on many tasks it also draws fresh private inputs each time (Section~\ref{sec:grading}), so a program whose error sits near the tolerance can pass one draw and fail the next. A single run's Score@1 therefore carries about 1.5 points of measurement uncertainty, and independent runs vary more (standard deviation 2.5--4.4 points), so differences of a few points between single runs do not order agents. On the machine that scores Table~\ref{tab:system_main}, judging the 64 final programs of GPT-6 Astra's first two runs twice gives a timing-only standard deviation of 1.4 and 1.6 points.

\subsection{What a Single Run Resolves}\label{sec:v11_resolution}
Table~\ref{tab:system_main} reports one run per agent and task. To see which of its differences exceed the variation across tasks, we resample the 32 tasks 10,000 times (seed 20261116) and compute paired 95\% percentile intervals for the Score@1 difference between two agents. Claude Sonnet 5.5 leads every non-Claude agent by at least \VSonnetLeadLo{} points at the lower end of its interval; it leads Claude Opus 5.5 by \VSonOpus{} points (\VSonOpusLo--\VSonOpusHi), and Claude Opus 5.5 leads GPT-6.1 Sol by \VOpusLead{} points (\VOpusSolLo--\VOpusSolHi). Below the two Claude agents, no two neighbors in the table have an interval that excludes zero; of all \VPairsTotal{} pairs of agents, \VPairsResolved{} do. These intervals cover task variation only. Re-judging fixed programs and running an agent again add the variation measured in Appendix~\ref{sec:v11_variance}, so differences of a few points between single runs do not order agents.

The order does not rest on the three workflows we wrote around a competition's optimization theme (\taskname{Paged LM generation}, \taskname{Tucker reconstruction} and \taskname{Graph sparsification}; Appendix~\ref{sec:taskorigins}). Recomputing Score@1 on the other \VSubN{} tasks gives a Kendall's $\tau_b$ of \VSubTau{} with the order of Table~\ref{tab:system_main}, and the two Claude agents stay at least \VSubLead{} points ahead of every other agent. Also leaving out \taskname{HPCG-derived}, whose operator admits a direct solver, gives $\tau_b=\VSubTauH$ and a lead of at least \VSubLeadH{} points.
\endgroup

\begingroup\color{revblue}
\section{Benchmark v1.0 Results and Run Selection}\label{sec:v10_results}
Table~\ref{tab:system_main_v10} reports the twelve agents evaluated on benchmark v1.0. Appendix~\ref{sec:eval_protocol} describes its agents and execution, and the per-task results, scoring sensitivity and analyses in Appendices~\ref{sec:diagnostic_metrics}--\ref{sec:trajectory_atlas} use these v1.0 runs. Section~\ref{sec:results} compares agents on v1.1 only: four of the twelve v1.0 agents, including GPT-6 Astra, ran once per task, and the variation below shows that one run does not order agents a few points apart. Scores are not compared across versions: besides the revised statements and verifiers, v1.1 changes the evaluation platform, the harness of Grok 4.6 and Gemini-3.8-Flash, and the reasoning effort of Kimi K3 and GLM-5.3 (Appendix~\ref{sec:v11_changes}). On v1.1, Hy3 runs under the same protocol as the other agents, and DeepSeek-V4.1-Flash is evaluated in its 1017 release (DeepSeek-V4.1-Flash-1017).

\begin{table}[!ht]
\centering\small\setlength{\tabcolsep}{3pt}
\caption{\textbf{\rev{Benchmark v1.0:} performance and recorded usage on 32 tasks} \rev{(one reported run per task: the median-reward run of three for eight agents and the only run for the other four; Appendix~\ref{sec:v10_selection})}. Scores and rates are percentages; GM is in $\times$. Effort denotes the requested reasoning setting. Steps and tokens are means per task; tokens include repeated and cached input. Usage is not reported for GPT-6 Astra, Gemini-3.8-Flash, and Grok 4.6 (---). Bold marks performance maxima.}
\label{tab:system_main_v10}
\begin{tabular*}{\linewidth}{@{\extracolsep{\fill}}llrrrrrr@{}}
\toprule
Model & Effort & Score@1 & GM & $A_{1.25}$ & $A_2$ & Steps & \shortstack[r]{Tokens\\(M)}\\
\midrule
\raisebox{-0.15ex}{\includegraphics[width=1em,height=1em,keepaspectratio]{figures/model-logos/openai.pdf}}\hspace{0.35em}GPT-6 Astra & high & \textbf{40.3} & 2.11 & \textbf{68.8} & \textbf{43.8} & \textemdash & \textemdash \\
\raisebox{-0.15ex}{\includegraphics[width=1em,height=1em,keepaspectratio]{figures/model-logos/hunyuan-color.pdf}}\hspace{0.35em}Hy4-Preview & high & 36.5 & \textbf{2.13} & 56.2 & 37.5 & 538.0 & 49.48 \\
\raisebox{-0.15ex}{\includegraphics[width=1em,height=1em,keepaspectratio]{figures/model-logos/deepseek-color.pdf}}\hspace{0.35em}DeepSeek-V4.1-Flash & high & 29.9 & 1.78 & 50.0 & 31.2 & 676.4 & 63.35 \\
\raisebox{-0.15ex}{\includegraphics[width=1em,height=1em,keepaspectratio]{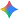}}\hspace{0.35em}Gemini-3.8-Flash & high & 29.9 & 1.71 & 50.0 & 25.0 & \textemdash & \textemdash \\
\raisebox{-0.15ex}{\includegraphics[width=1em,height=1em,keepaspectratio]{figures/model-logos/zai.pdf}}\hspace{0.35em}GLM-5.3 & max & 28.1 & 1.77 & 40.6 & 21.9 & 967.1 & 92.89 \\
\raisebox{-0.15ex}{\includegraphics[width=1em,height=1em,keepaspectratio]{figures/model-logos/zai.pdf}}\hspace{0.35em}GLM-5.3-Flash & high & 27.3 & 1.65 & 43.8 & 21.9 & 773.8 & 74.98 \\
\raisebox{-0.15ex}{\includegraphics[width=1em,height=1em,keepaspectratio]{figures/model-logos/grok.pdf}}\hspace{0.35em}Grok 4.6 & high & 26.2 & 1.66 & 43.8 & 21.9 & \textemdash & \textemdash \\
\raisebox{-0.15ex}{\includegraphics[width=1em,height=1em,keepaspectratio]{figures/model-logos/deepseek-color.pdf}}\hspace{0.35em}DeepSeek-V4-Pro & high & 20.6 & 1.46 & 43.8 & 12.5 & 471.7 & 46.17 \\
\raisebox{-0.15ex}{\includegraphics[width=1em,height=1em,keepaspectratio]{figures/model-logos/zai.pdf}}\hspace{0.35em}GLM-5.2 & high & 19.9 & 1.46 & 34.4 & 15.6 & 529.8 & 45.61 \\
\raisebox{-0.15ex}{\includegraphics[width=1em,height=1em,keepaspectratio]{figures/model-logos/deepseek-color.pdf}}\hspace{0.35em}DeepSeek-V4-Flash & high & 19.1 & 1.44 & 34.4 & 12.5 & 404.0 & 38.70 \\
\raisebox{-0.15ex}{\includegraphics[width=1em,height=1em,keepaspectratio]{figures/model-logos/kimi.pdf}}\hspace{0.35em}Kimi K3 & max & 19.0 & 1.34 & 34.4 & 15.6 & 433.0 & 37.37 \\
\raisebox{-0.15ex}{\includegraphics[width=1em,height=1em,keepaspectratio]{figures/model-logos/hunyuan-color.pdf}}\hspace{0.35em}Hy3 & high & 9.8 & 1.25 & 12.5 & 6.2 & 254.3 & 20.92 \\
\bottomrule
\end{tabular*}
\end{table}

\subsection{Reported Run}\label{sec:v10_selection}
Eight agents ran three times per task: Hy4-Preview, DeepSeek-V4.1-Flash, GLM-5.3-Flash, Grok 4.6, DeepSeek-V4-Pro, GLM-5.2, DeepSeek-V4-Flash and Hy3. Their three runs are the three most recent eligible runs by completion time, and Table~\ref{tab:system_main_v10} reports the run with the median reward, breaking ties by completion time and then trial identifier; correctness and usage play no part in the choice. Because of cost, GPT-6 Astra, Gemini-3.8-Flash, GLM-5.3 and Kimi K3 ran once per task, so their Score@1 is the reward of that single run.

\subsection{Variation Across Three Runs}\label{sec:v10_repeats}
Table~\ref{tab:v10_repeats} shows how much Score@1 changes when the same agent runs again. Round $i$ consists of the $i$-th run of every task by completion time. Across rounds, Score@1 varies by 0.6--3.9 points (standard deviation), although the reward of the same agent on the same task varies by 6.9--13.2 points on average. The ranking changes little: Hy4-Preview is first, DeepSeek-V4.1-Flash second and Hy3 last in every round, and the only changes are swaps between Grok 4.6 and GLM-5.3-Flash (third and fourth) and among DeepSeek-V4-Pro, DeepSeek-V4-Flash and GLM-5.2 (fifth to seventh); Kendall's $\tau$ between two rounds is 0.79--0.86.

\begin{table}[!ht]
\centering\small
\caption{\textbf{Score@1 (\%) of the v1.0 agents with three runs per task, by round.} ``Reported'' is the run of Table~\ref{tab:system_main_v10}, the median-reward run of each task.}
\label{tab:v10_repeats}
\begin{tabular*}{\linewidth}{@{\extracolsep{\fill}}lrrrrrr@{}}
\toprule
Agent & Round 1 & Round 2 & Round 3 & Mean & SD & Reported\\
\midrule
Hy4-Preview & 34.61 & 32.74 & 40.18 & 35.84 & 3.87 & 36.46\\
DeepSeek-V4.1-Flash & 27.87 & 29.27 & 32.07 & 29.74 & 2.14 & 29.94\\
Grok 4.6 & 27.53 & 26.03 & 24.93 & 26.16 & 1.30 & 26.24\\
GLM-5.3-Flash & 25.42 & 22.24 & 29.07 & 25.58 & 3.41 & 27.34\\
DeepSeek-V4-Flash & 19.23 & 20.07 & 22.84 & 20.72 & 1.89 & 19.08\\
DeepSeek-V4-Pro & 19.47 & 19.78 & 21.70 & 20.32 & 1.21 & 20.63\\
GLM-5.2 & 19.29 & 19.92 & 14.22 & 17.81 & 3.12 & 19.92\\
Hy3 & 11.16 & 12.05 & 12.17 & 11.80 & 0.55 & 9.76\\
\bottomrule
\end{tabular*}
\end{table}

\subsection{Run Accounting}\label{sec:v10_accounting}
No agent--task slot is missing or scored as zero for an infrastructure reason. Interrupted runs receive no score, and a continuation or a new run fills the slot; failures caused by the agent, including incorrect output, out-of-memory termination (exit code~137) and timeout, are scored. Table~\ref{tab:v10_accounting} accounts for the runs of the eight three-run agents. The four one-run agents have exactly one scored record per task; for Kimi K3 on 16 tasks and GLM-5.3 on one, that record is a continuation run and its parent is not scored.

\begin{table}[!ht]
\centering\small
\caption{\textbf{Run accounting for the eight v1.0 agents with three runs per task (32 tasks).}}
\label{tab:v10_accounting}
\begin{tabular*}{\linewidth}{@{\extracolsep{\fill}}lrl@{}}
\toprule
Category & Runs & Scored\\
\midrule
Scored panel ($8\times32\times3$) & 768 & yes\\
API failure after exhausting retries & 215 & no; continuation or new run fills the slot\\
Infrastructure cancellation & 108 & no; as above\\
Infrastructure runtime error & 18 & no; as above\\
Ended with an indeterminate error & 4 & no; as above\\
Verifier result missing or verifier fault & 4 & no\\
Eligible but not among the three most recent & 65 (in 49 groups) & no\\
Parent replaced by its continuation & 13 & no; the continuation fills the slot\\
\bottomrule
\end{tabular*}
\end{table}

Out-of-memory terminations are scored as the agent's failure; the only such run left unscored lacks a verifier result. Which three runs enter the panel depends only on completion time and run status, never on scores; when to schedule a replacement depended on coverage and on API availability. A continuation inherits its parent's session history and has the same eight-hour timeout, not the parent's remaining time, so parent and continuation together can exceed eight hours: in an earlier 842-run snapshot, the 91 continuations and their parents took a median of 2.92 hours together, and 10 exceeded eight hours.
\endgroup

\section{Appendix Overview and Evidence Sources}\label{sec:evidence}
\rev{Appendices~\ref{sec:v11_changes} and~\ref{sec:v10_results} describe benchmark v1.1 and the v1.0 results; the other} appendices follow the order of the main text (Table~\ref{tab:appendix_map}). Each numerical result is computed from one of the frozen snapshots listed in Table~\ref{tab:evidence_index}.

\begin{table}[!htbp]
\centering\small
\caption{\textbf{Where the appendices support the main text.}}
\label{tab:appendix_map}
\begin{tabularx}{\linewidth}{@{}>{\raggedright\arraybackslash}p{.3\linewidth}>{\raggedright\arraybackslash}X@{}}
\toprule
Main text & Appendix\\
\midrule
Benchmark construction (\S\ref{sec:benchmark}) & Tasks, sources, resources and contracts (\ref{sec:inventory}); verifier validation controls (\ref{sec:validation})\\
Evaluation metrics (\S\ref{sec:protocol}) & Agents, eligibility, usage (\ref{sec:eval_protocol}); timing calibration, score sensitivity (\ref{sec:diagnostic_metrics})\\
Main results (\S\ref{sec:results}) & \rev{Benchmark v1.1 changes, thresholds, configuration baseline, timing noise and what a single run resolves (\ref{sec:v11_changes}); v1.0 results and run selection (\ref{sec:v10_results});} per-task rewards of the \rev{v1.0} reported runs (\ref{sec:taskresults})\\
Analysis sample and labels (\S\ref{sec:analysis}) & Annotation procedures and their audits (\ref{sec:annotation_methods})\\
Changes, candidates (\S\ref{sec:rq1}--\S\ref{sec:rq2}) & Technique and workflow analysis (\ref{sec:optimization_panel})\\
Interacting choices (\S\ref{sec:mechanism_controls}) & Dependency review (\ref{sec:trajectory_annotations}); controlled experiments (\ref{sec:controlled_replay}); case records (\ref{sec:mechanism_evidence})\\
All trajectories & Action sequences and outcomes of the 480 runs (\ref{sec:trajectory_atlas})\\
\bottomrule
\end{tabularx}
\end{table}

\begin{table}[!htbp]
\centering\small
\caption{\textbf{Evidence sources.} Each row is a frozen snapshot; the earlier snapshot supports only the timing and aggregation checks named here.}
\label{tab:evidence_index}
\begin{tabularx}{\linewidth}{@{}>{\raggedright\arraybackslash}p{.4\linewidth}>{\raggedright\arraybackslash}X@{}}
\toprule
Snapshot & Used for\\
\midrule
\rev{Benchmark v1.1: first run of every agent; three runs of GPT-6 Astra and Hy4-Preview; fixed configuration-only baseline; repeated judgments of fixed submissions} & \rev{Table~\ref{tab:system_main}, Figures~\ref{fig:dataset_diversity} and~\ref{fig:acceleration_distribution}; Appendix~\ref{sec:v11_changes}}\\
\rev{Benchmark v1.0} main evaluation: 12 agents, 384 runs & Table~\rev{\ref{tab:system_main_v10}}, Table~\ref{tab:task_results}; \rev{Appendix~\ref{sec:v10_results};} scoring sensitivity (\ref{sec:reference_sensitivity})\\
Analysis: 480 runs of five agents, three per task, with trajectories and public-test records & Section~\ref{sec:analysis}; Appendices~\ref{sec:annotation_methods}--\ref{sec:trajectory_annotations}, \ref{sec:mechanism_evidence} and~\ref{sec:trajectory_atlas}\\
Earlier 842-run snapshot, per-case timings & Baseline perturbation (\ref{sec:timing_calibration}); workload aggregation check (\ref{sec:workload_aggregation})\\
Controlled replays of fixed programs & Appendices~\ref{sec:validation}, \ref{sec:diagnostic_metrics} and~\ref{sec:controlled_replay}; conditions are stated per experiment\\
\bottomrule
\end{tabularx}
\end{table}

\paragraph{Terms.}
A \emph{run} (also called an attempt) is one agent's optimization session on one task. Its \emph{trajectory} records the agent's messages, tool calls and outputs. A \emph{candidate} is a program the agent sends to the public tests during the run; the \emph{final submission} is the frozen program the verifier rebuilds and scores.

\paragraph{Conventions for interpreting the analyses.}
Comparisons between runs are observational. Where stated, they hold the model and task fixed. Other changes still co-occur within a run, so these comparisons measure associations rather than the effect of a single technique or behavior. Unless stated otherwise, intervals are pointwise 95\% percentile intervals obtained by resampling the 32 tasks, keeping all runs of a task together. They are conditional on the observed runs and are not adjusted for multiple comparisons. Controlled replays measure the named programs under the stated workloads, resources and settings, and their timings are reported separately from the public-test feedback that the agent saw.

\section{Tasks, Resources, and Contracts}\label{sec:inventory}\label{sec:construction_details}
\paragraph{Task definition.}
A task $t$ specifies an initial program $a_t^0$, a permitted edit space $\mathcal{A}_t$, a public test $P_t$, a verifier $V_t$, and an interaction budget $b_t$. The task instruction defines the required computation, outputs and editable paths. During a run, the agent produces candidates $a_t^1,a_t^2,\ldots\in\mathcal{A}_t$, guided by code inspection, its own executions and the public responses $P_t(a_t^k)$, and leaves a final submission $\hat a_t$. The verifier scores $\hat a_t$ on private inputs; the agent does not see this reward during the run. The score therefore evaluates the submitted program under the task contract, independent of what the agent claims about it.

\paragraph{Resources.}
Each task package fixes its CPU cores, memory, agent interaction limit and verifier time limit. Table~\ref{tab:resources} lists these allocations and the task-package version, with memory in GiB, interaction limits in hours and verifier limits in minutes.
\begin{longtable}{@{}p{.36\linewidth}rrrrr@{}}
\caption{\textbf{Resource allocations and task-package versions.}}\label{tab:resources}\\
\toprule Task & Version & CPUs & GiB & Agent h & Verify min\\\midrule\endfirsthead
\toprule Task & Version & CPUs & GiB & Agent h & Verify min\\\midrule\endhead
\taskname{GROMACS} & 12.0.3 & 4 & 6 & 8 & 390 \\
\taskname{LAMMPS} & 4.1.3 & 4 & 8 & 8 & 30 \\
\taskname{GPAW} & 1.0.2 & 4 & 6 & 8 & 233.3 \\
\taskname{Quantum ESPRESSO} & 1.2.2 & 4 & 6 & 8 & 233.3 \\
\taskname{SIESTA} & 1.2.2 & 4 & 6 & 8 & 233.3 \\
\taskname{Fluidity} & 1.0.2 & 4 & 8 & 8 & 233.3 \\
\taskname{ICON} & 1.0.2 & 4 & 6 & 8 & 200 \\
\taskname{MPAS-A} & 1.0.2 & 4 & 6 & 8 & 233.3 \\
\taskname{NEMO} & 1.0.2 & 4 & 8 & 8 & 233.3 \\
\taskname{OpenFOAM} & 1.0.3 & 4 & 8 & 8 & 40 \\
\taskname{Palabos} & 2.0.2 & 4 & 6 & 8 & 120 \\
\taskname{SU2} & 2.0.2 & 4 & 8 & 8 & 60 \\
\taskname{OpenMC} & 3.0.4 & 4 & 6 & 8 & 120 \\
\taskname{ShengBTE} & 1.0.2 & 4 & 6 & 8 & 233.3 \\
\taskname{VPIC} & 1.2.2 & 4 & 6 & 8 & 150 \\
\taskname{PRESTO} & 2.0.2 & 4 & 6 & 8 & 120 \\
\taskname{ABySS} & 1.0.3 & 4 & 8 & 8 & 30 \\
\taskname{MHM2} & 1.0.2 & 4 & 6 & 8 & 150 \\
\taskname{MrBayes} & 2.0.2 & 4 & 6 & 8 & 120 \\
\taskname{wtdbg2} & 2.0.3 & 4 & 8 & 8 & 30 \\
\taskname{HPCG} & 12.0.3 & 4 & 6 & 8 & 120 \\
\taskname{HPL} & 11.0.2 & 4 & 6 & 8 & 120 \\
\taskname{Graph sparsification} & 3.0.2 & 4 & 4 & 8 & 120 \\
\taskname{miniVite} & 11.0.2 & 4 & 6 & 8 & 120 \\
\taskname{NUFFT Imaging} & 1.0.2 & 4 & 6 & 8 & 200 \\
\taskname{Tucker reconstruction} & 3.1.6 & 8 & 16 & 8 & 120 \\
\taskname{QuEST} & 3.0.5 & 4 & 8 & 8 & 60 \\
\taskname{QiboTN} & 3.2.7 & 8 & 32 & 8 & 833.3 \\
\taskname{Paged LM generation} & 1.0.2 & 4 & 6 & 8 & 200 \\
\taskname{ThunderSVM} & 4.3.4 & 4 & 8 & 8 & 60 \\
\taskname{DataLife} & 1.0.2 & 4 & 6 & 8 & 233.3 \\
\taskname{IO500} & 12.0.2 & 4 & 6 & 8 & 120 \\

\bottomrule
\end{longtable}

\begin{table}[!ht]
\caption{\textbf{Task collection by application area.} All 32 tasks carry equal weight in every aggregate.}
\label{tab:coverage}
\centering\small
\begin{tabularx}{\linewidth}{@{}>{\raggedright\arraybackslash}p{.34\linewidth}rY@{}}
\toprule
Application area & $N$ & Tasks\\
\midrule
Molecular simulation / electronic structure & 5 & \taskname{GROMACS}, \taskname{LAMMPS}, \taskname{GPAW}, \taskname{Quantum ESPRESSO}, \taskname{SIESTA}\\
Fluids / atmosphere / ocean & 7 & \taskname{Fluidity}, \taskname{ICON}, \taskname{MPAS-A}, \taskname{NEMO}, \taskname{OpenFOAM}, \taskname{Palabos}, \taskname{SU2}\\
Transport / plasma / astronomy & 4 & \taskname{OpenMC}, \taskname{ShengBTE}, \taskname{VPIC}, \taskname{PRESTO}\\
Bioinformatics / phylogenetics & 4 & \taskname{ABySS}, \taskname{MHM2}, \taskname{MrBayes}, \taskname{wtdbg2}\\
Numerics / tensors / graphs & 6 & \taskname{HPCG-derived}, \taskname{HPL-derived}, \taskname{NUFFT imaging}, \taskname{Tucker reconstruction}, \taskname{Graph sparsification}, \taskname{miniVite}\\
Quantum circuits & 2 & \taskname{QuEST}, \taskname{QiboTN}\\
Machine learning & 2 & \taskname{Paged LM generation}, \taskname{ThunderSVM}\\
Scientific I/O & 2 & \taskname{DataLife}, \taskname{IO500-derived}\\
\midrule
Total & 32 & \\
\bottomrule
\end{tabularx}
\end{table}

\paragraph{Computational contracts.}
Each contract names what the agent may edit, what is timed and what the program must produce. Table~\ref{tab:contracts} gives three representative contracts, and every task package states these three elements for its own workflow.

\begin{table}[!ht]
\caption{\textbf{Example computational contracts.} The \taskname{HPCG-derived} task is an application optimization problem built on the \taskname{HPCG} source, not the official \taskname{HPCG} competition benchmark.}
\label{tab:contracts}
\centering\footnotesize\setlength{\tabcolsep}{3pt}
\begin{tabularx}{\linewidth}{@{}>{\raggedright\arraybackslash}p{.17\linewidth}YYY@{}}
\toprule
Task & Permitted edits & Timed execution & Required outputs and checks\\
\midrule
\taskname{GROMACS} & Full source tree and supported build settings & Complete \texttt{mdrun}; excludes build and input preparation & Prescribed physical calculation, trajectories, energies, final state, deterministic replay\\
\addlinespace[3pt]
\taskname{HPCG-derived} & Source tree, solver algorithm, entry point, build settings & Input, matrix construction, setup, two solves, output & Two full solutions; absolute error $\leq10^{-8}$, relative residual $\leq10^{-9}$; identical repeated vectors\\
\addlinespace[3pt]
\taskname{Paged generation} & Application modules and helpers & Model loading, prefill, decoding, checkpoint, fresh-process resume & Full-model states, KV caches, sampling probabilities and tokens; numerical tolerances and exact recovery outputs\\
\bottomrule
\end{tabularx}
\end{table}

\subsection{Task sources and reconstruction}\label{sec:taskorigins}
\paragraph{Competition provenance.}
Table~\ref{tab:competition_sources} assigns each task to the competition family recorded in its construction history, which identifies the application or optimization theme that motivated its reconstruction. An application that also appears in other competitions is counted once, under its recorded source.

\begin{table}[H]
\caption{Recorded competition sources for all 32 tasks. Names link to organizer or event records.}
\label{tab:competition_sources}
\centering\small
\begin{tabularx}{\linewidth}{@{}p{.43\linewidth}rY@{}}
\toprule
Competition & $N$ & Tasks\\
\midrule
\href{https://asc-events.net/StudentChallenge/index.html}{Asia Supercomputer Community (ASC) Student Supercomputer Challenge} & 14 & \taskname{ABySS}, \taskname{QiboTN}, \taskname{Fluidity}, \taskname{HPCG}, \taskname{HPL}, \taskname{NEMO}, \taskname{OpenFOAM}, \taskname{Palabos}, \taskname{PRESTO}, \taskname{QuEST}, \taskname{ShengBTE}, \taskname{Siesta}, \taskname{SU2}, \taskname{wtdbg2}\\
\addlinespace[3pt]
\href{https://sc26.supercomputing.org/students/cluster-competition-history/}{Student Cluster Competition at the International Conference for High Performance Computing, Networking, Storage, and Analysis (SC)} & 10 & \taskname{DataLife}, \taskname{GROMACS}, \taskname{ICON}, \taskname{IO500}, \taskname{LAMMPS}, \taskname{MPAS-A}, \taskname{MrBayes}, \taskname{OpenMC}, \taskname{Quantum ESPRESSO}, \taskname{VPIC}\\
\addlinespace[3pt]
\href{https://paraac.com/events/pac}{Parallel Application Challenge} & 4 & \taskname{NUFFT}, \taskname{Paged LM generation}, \taskname{Tensor}, \taskname{ThunderSVM}\\
\addlinespace[3pt]
\href{https://isc-hpc.com/program/student-cluster-competition/}{Student Cluster Competition at the International Supercomputing Conference (ISC)} & 2 & \taskname{GPAW}, \taskname{MHM2}\\
\addlinespace[3pt]
\href{https://www.sc22.supercomputing.org/program/studentssc/indyscc/index.htm}{IndySCC, the remote Student Cluster Competition at SC} & 1 & \taskname{miniVite}\\
\addlinespace[3pt]
\href{https://cnic.cas.cn/gzdt/202111/t20211126_6272303.html}{ACM-China International Parallel Computing Challenge} & 1 & \taskname{Graph sparsification}\\
\addlinespace[3pt]
\bottomrule
\end{tabularx}
\end{table}

\paragraph{From source material to a runnable task.}
Competition descriptions provided application names and optimization themes; upstream repositories and examples supplied reusable code and scientific inputs. We adapted these materials into CPU workloads with explicit build and execution entry points, collected outputs, and permitted edit spaces. Some tasks extend earlier component tests into broader application workflows. For example, \taskname{HPL} now performs an MPI dense solve and exports the full solution, and \taskname{wtdbg2} performs graph assembly and consensus. Where the package specifies it, the earlier component tests remain as regression checks. Where original competition inputs were missing, we used upstream examples, public datasets, or generated inputs meeting the requirements.

\paragraph{Source scope and workload scope.}
A complete source tree does not mean the workload exercises every application feature. For example, the full \taskname{miniVite} program runs the first distributed Louvain phase, \taskname{GPAW} performs two geometry steps, and \taskname{ShengBTE} covers bulk transport. Other tasks expose selected source components. \taskname{OpenFOAM} permits changes to its \texttt{pisoFoam} solver and case over fixed system libraries, \taskname{SU2} submissions contain specified solver subtrees and a case, and \taskname{DataLife} permits changes to monitoring and analysis around a fixed I/O driver. These boundaries determine which implementation choices an agent can change.

Some workloads were built from an optimization theme rather than reconstructed from the original competition program. \taskname{Paged generation} runs SmolLM2 with task-authored NumPy code, the \taskname{tensor} task combines \taskname{TAPP} with a Tucker compression and reconstruction driver, and the graph task independently implements a feGRASS-inspired sparsification and solver pipeline. Their contracts specify these computations, which differ from the original PAC or IPCC payloads. Table~\ref{tab:taskorigins} records the resulting source and workflow choices. The versioned task packages define the evaluated collection, which adapts the competition workloads rather than reproducing them.

Source identities come from the construction records. Ordinary-file comparisons against upstream independently checked the \taskname{QuEST}, \taskname{QiboTN}, \taskname{TAPP}, and \taskname{Palabos} sources; \taskname{QiboTN} includes a CPU backend adaptation. The table marks unresolved commits and archive mappings explicitly.

\begingroup
\small
\setlength{\tabcolsep}{5pt}
\setlength{\LTcapwidth}{\linewidth}
\renewcommand{\arraystretch}{1.06}
\begin{longtable}{@{}>{\raggedright\arraybackslash}p{.27\linewidth}>{\raggedright\arraybackslash}p{.22\linewidth}>{\raggedright\arraybackslash}p{\dimexpr.51\linewidth-20pt\relax}@{}}
\caption{\textbf{Task origins and reconstructed workflows.} Repository names link to GitHub unless another host is named; $\dagger$ marks a source mirror. Releases are upstream versions, distinct from task-package versions; commits are abbreviated.}\label{tab:taskorigins}\\
\toprule
Task / source repository & Release / revision & Supplied code and retained computation\\
\midrule
\endfirsthead
\multicolumn{3}{@{}l}{\small\textit{Table~\ref{tab:taskorigins} continued.}}\\[3pt]
\toprule
Task / source repository & Release / revision & Supplied code and retained computation\\
\midrule
\endhead
\midrule
\multicolumn{3}{r@{}}{\small\textit{Continued on the next page.}}\\
\endfoot
\bottomrule
\endlastfoot
\textbf{\taskname{ABySS}}\newline\href{https://github.com/BirolLab/abyss}{\textcolor{draftblue}{\texttt{BirolLab/\allowbreak abyss}}} & 1.9.0\newline\texttt{07629837046e} & Full application source; generated short reads drive assembly, graph/coverage outputs and contig statistics.\\
\addlinespace[4pt]
\textbf{\taskname{QiboTN}}\newline\href{https://github.com/qiboteam/qibotn}{\textcolor{draftblue}{\texttt{qiboteam/\allowbreak qibotn}}} & \taskname{QiboTN} 0.0.6\newline\texttt{283bab987e89} & Full \taskname{QiboTN} tree with a disclosed CPU backend patch; task-authored circuits produce complete states and observables.\\
\addlinespace[4pt]
\textbf{\taskname{DataLife}}\newline\href{https://github.com/pnnl/datalife}{\textcolor{draftblue}{\texttt{pnnl/\allowbreak datalife}}} & Release name unknown\newline\texttt{ee9d3939a323} & FlowMonitor/FlowAnalysis/tools source; monitoring and DAG analysis of a fixed I/O driver. Separate HDF5/DaYu projects excluded.\\
\addlinespace[4pt]
\textbf{\taskname{Fluidity}}\newline\href{https://github.com/FluidityProject/fluidity}{\textcolor{draftblue}{\texttt{FluidityProject/\allowbreak fluidity}}} & 4.1.16\newline\texttt{4fc02e3da1cb} & Full source; selected MPI redistribution and DG heat workflows with field outputs and checkpoint recovery.\\
\addlinespace[4pt]
\textbf{\taskname{GPAW}}\newline\href{https://gitlab.com/gpaw/gpaw}{\textcolor{draftblue}{GitLab: gpaw/\allowbreak gpaw}} & 21.1.0 sdist\newline\texttt{c5b9c0c91bf0}\newline Historical tag; sdist mapping unverified & Full release sdist and upstream PAW setups; SCF, two BFGS steps and restart. These steps do not establish geometry convergence.\\
\addlinespace[4pt]
\textbf{\taskname{GROMACS}}\newline\href{https://github.com/gromacs/gromacs}{\textcolor{draftblue}{\texttt{gromacs/\allowbreak gromacs}}}$^{\dagger}$ & 2020.4\newline Commit unresolved & Full source with build/numerical preparation patches; prescribed water/argon MD trajectories and energies.\\
\addlinespace[4pt]
\textbf{\taskname{HPCG-derived}}\newline\href{https://github.com/hpcg-benchmark/hpcg}{\textcolor{draftblue}{\texttt{hpcg-\allowbreak benchmark/\allowbreak hpcg}}} & 3.1\newline\texttt{e3fe052b572b} & Full source plus an authored entry point; a prescribed matrix with two right-hand sides may use replacement solvers. Official \taskname{HPCG} compliance is not claimed.\\
\addlinespace[4pt]
\textbf{\taskname{HPL-derived}}\newline\href{https://github.com/icl-utk-edu/hpl}{\textcolor{draftblue}{\texttt{icl-\allowbreak utk-\allowbreak edu/\allowbreak hpl}}} & 2.3\newline Commit unresolved & Full source plus matrix-input and solution export; bounded MPI dense solves, not a TOP500 run.\\
\addlinespace[4pt]
\textbf{\taskname{ICON}}\newline\href{https://github.com/spcl/icon-dace}{\textcolor{draftblue}{\texttt{spcl/\allowbreak icon-\allowbreak dace}}}$^{\dagger}$ & 2024.10 public\newline\texttt{a1324166fa60} & Full recorded source mirror; selected Hadley/JABW atmosphere cases on an upstream global grid, with conservation and restart outputs.\\
\addlinespace[4pt]
\textbf{\taskname{IO500-derived}}\newline\href{https://github.com/IO500/io500}{\textcolor{draftblue}{\texttt{IO500/\allowbreak io500}}} & \taskname{IO500} sc20\newline\texttt{65b0397101a8} & Full IO500/IOR/mdtest/pfind sources; all 12 phases, shortened below the official duration rule.\\
\addlinespace[4pt]
\textbf{\taskname{feGRASS-derived}}\newline\href{https://doi.org/10.1109/TCAD.2021.3060647}{\textcolor{draftblue}{\taskname{feGRASS} paper}} & Independent implementation\newline No author-code import & Independent six-file implementation; generated graph sparsification, preconditioning and multi-RHS PCG. Original authors' source and IPCC payload are not used.\\
\addlinespace[4pt]
\textbf{\taskname{LAMMPS}}\newline\href{https://github.com/lammps/lammps}{\textcolor{draftblue}{\texttt{lammps/\allowbreak lammps}}} & patch\_23Oct2017\newline\texttt{39df9f5d9467} & Full serial application source; generated MD inputs, trajectories and restart continuation. Partial reconstruction of the competition workload.\\
\addlinespace[4pt]
\textbf{\taskname{MHM2}}\newline\href{https://bitbucket.org/berkeleylab/mhm2}{\textcolor{draftblue}{Bitbucket: berkeleylab/\allowbreak mhm2}} & 2.0.0 release\newline Commit unresolved & Full release and recorded dependencies; generated two-genome, two-round coassembly with graphs, alignments and abundance outputs; UPC++ smp.\\
\addlinespace[4pt]
\textbf{\taskname{miniVite}}\newline\href{https://github.com/ECP-ExaGraph/miniVite}{\textcolor{draftblue}{\texttt{ECP-\allowbreak ExaGraph/\allowbreak miniVite}}} & Release name unknown\newline\texttt{4cbf5beb58c6} & Full \taskname{miniVite} source with partition-output additions; distributed Louvain first phase, not the multi-phase Vite application.\\
\addlinespace[4pt]
\textbf{\taskname{MPAS atmosphere}}\newline\href{https://github.com/MPAS-Dev/MPAS-Model}{\textcolor{draftblue}{\texttt{MPAS-\allowbreak Dev/\allowbreak MPAS-\allowbreak Model}}} & 8.4.1\newline\texttt{91c5eac175ee} & Full application and pinned physics dependencies; upstream 240-km grid, bounded forecast and restart continuation.\\
\addlinespace[4pt]
\textbf{\taskname{MrBayes}}\newline\href{https://github.com/NBISweden/MrBayes}{\textcolor{draftblue}{\texttt{NBISweden/\allowbreak MrBayes}}} & 3.2.7a\newline\texttt{8819c865c312} & Full source; upstream and generated alignments, two four-chain runs, midpoint recovery and posterior summaries.\\
\addlinespace[4pt]
\textbf{\taskname{NEMO}}\newline\href{https://forge.nemo-ocean.eu/nemo/nemo}{\textcolor{draftblue}{\taskname{NEMO} forge: nemo/\allowbreak nemo}} & 5.0.2\newline\texttt{dcc7fb8c1779} & Full source; LOCK\_EXCHANGE ocean run with full fields and checkpoint continuation.\\
\addlinespace[4pt]
\textbf{\taskname{OpenFOAM}}\newline\href{https://www.openfoam.com/news/main-news/openfoam-v1912}{\textcolor{draftblue}{\taskname{OpenFOAM} release}} & Debian 1912.200626\newline Commit unresolved & Editable pisoFoam solver and cavity case, plus legacy LDU code; fixed Debian \taskname{OpenFOAM} libraries. Transient full-field CFD.\\
\addlinespace[4pt]
\textbf{\taskname{OpenMC}}\newline\href{https://github.com/openmc-dev/openmc}{\textcolor{draftblue}{\texttt{openmc-\allowbreak dev/\allowbreak openmc}}} & 0.16.0\newline\texttt{617d35a5063c} & Full source; upstream C5G7 cross-section constants in task-authored fuel-pin geometry, with full tallies and source-bank recovery.\\
\addlinespace[4pt]
\textbf{\taskname{NUFFT imaging}}\newline\href{https://github.com/flatironinstitute/finufft}{\textcolor{draftblue}{\texttt{flatironinstitute/\allowbreak finufft}}} & FINUFFT 2.4.1\newline\texttt{e7144a5c08cb} & Full FINUFFT source plus authored binning and inverse-imaging drivers; complete reconstruction and CG history.\\
\addlinespace[4pt]
\textbf{\taskname{Paged LM generation}}\newline\href{https://huggingface.co/HuggingFaceTB/SmolLM2-135M}{\textcolor{draftblue}{HF: HuggingFaceTB/\allowbreak SmolLM2-135M}} & SmolLM2-135M\newline\texttt{93efa2f097d5}\newline Model revision; driver commit unknown & Full SmolLM2 weights with authored NumPy code; prefill, sampling and resumed generation. No equivalence to the PAC DeepSeek indexer is shown.\\
\addlinespace[4pt]
\textbf{\taskname{Tucker reconstruction}}\newline\href{https://github.com/TAPPorg/reference-implementation}{\textcolor{draftblue}{\texttt{TAPPorg/\allowbreak reference-\allowbreak implementation}}} & \taskname{TAPP} 0.5.0\newline\texttt{77c32d744ee6} & Full \taskname{TAPP} source plus authored HOSVD/Tucker drivers and bridge; all reconstruction stages. Not the original large PAC contraction workload.\\
\addlinespace[4pt]
\textbf{\taskname{ThunderSVM}}\newline\href{https://github.com/Xtra-Computing/thundersvm}{\textcolor{draftblue}{\texttt{Xtra-\allowbreak Computing/\allowbreak thundersvm}}} & 0.3.4\newline\texttt{e71013a88fb5} & Full CPU application build/source surface; public a9a-derived inputs, C-SVC training, saved-model reload and prediction.\\
\addlinespace[4pt]
\textbf{\taskname{Palabos}}\newline\href{https://github.com/omalaspinas/palabos}{\textcolor{draftblue}{\texttt{omalaspinas/\allowbreak palabos}}}$^{\dagger}$ & 2.3.0\newline\texttt{bba6dc6464fa} & Full source plus authored driver/CMake; upstream cavity-derived and periodic shear simulations with complete field outputs.\\
\addlinespace[4pt]
\textbf{\taskname{PRESTO}}\newline\href{https://github.com/scottransom/presto}{\textcolor{draftblue}{\texttt{scottransom/\allowbreak presto}}} & 3.0.1\newline\texttt{7d569eff79ec} & Full source with disclosed planning/build adaptations; generated pulse time series drive FFT, acceleration search and folding.\\
\addlinespace[4pt]
\textbf{\taskname{Quantum ESPRESSO}}\newline\href{https://github.com/QEF/q-e}{\textcolor{draftblue}{\texttt{QEF/\allowbreak q-\allowbreak e}}} & 6.8 ReleasePack\newline\texttt{03a7fa640903} & Full release and pinned devxlib; selected SCF/NSCF, density, DOS and checkpoint workflows with upstream pseudopotentials.\\
\addlinespace[4pt]
\textbf{\taskname{QuEST}}\newline\href{https://github.com/QuEST-Kit/QuEST}{\textcolor{draftblue}{\texttt{QuEST-\allowbreak Kit/\allowbreak QuEST}}} & 4.2.0\newline\texttt{9d7618d7263e} & Full upstream source plus task runners; state-vector and density-matrix circuits with complete states and probabilities.\\
\addlinespace[4pt]
\textbf{\taskname{ShengBTE}}\newline\href{https://bitbucket.org/sousaw/shengbte}{\textcolor{draftblue}{Bitbucket: sousaw/\allowbreak shengbte}} & 1.5.1\newline\texttt{e9f6fdc3362f} & Full source and upstream force constants; bulk RTA and iterative BTE at prescribed settings. Nanowire workflows are outside the task.\\
\addlinespace[4pt]
\textbf{\taskname{SIESTA}}\newline\href{https://gitlab.com/siesta-project/siesta}{\textcolor{draftblue}{GitLab: siesta-project/\allowbreak siesta}} & 4.1.5\newline\texttt{c3221beccffa} & Full source and upstream tests/pseudopotentials; LDA calculations and density-matrix restart. The older matrix proxy is only retained regression coverage.\\
\addlinespace[4pt]
\textbf{\taskname{SU2}}\newline\href{https://github.com/su2code/SU2}{\textcolor{draftblue}{\texttt{su2code/\allowbreak SU2}}} & 8.5.0\newline\texttt{12eb826f049e} & Only Common/SU2\_CFD and the case are submitted over a fixed full build tree; serial NACA0012 Euler flow and five complete snapshots.\\
\addlinespace[4pt]
\textbf{\taskname{VPIC}}\newline\href{https://github.com/lanl/vpic}{\textcolor{draftblue}{\texttt{lanl/\allowbreak vpic}}} & 1.2\newline\texttt{f29f1545d339} & Full source; bounded upstream Harris reconnection case with complete diagnostic dumps and checkpoint recovery.\\
\addlinespace[4pt]
\textbf{\taskname{wtdbg2}}\newline\href{https://github.com/ruanjue/wtdbg2}{\textcolor{draftblue}{\texttt{ruanjue/\allowbreak wtdbg2}}} & 2.5\newline\texttt{b77c5657c809} & Full assembler and consensus source; generated noisy PacBio/ONT-style reads, graph/layout outputs and consensus. Original competition reads are not reproduced.\\
\addlinespace[4pt]
\end{longtable}
\endgroup

\section{Verifier Validation Controls}\label{sec:validation}
The verifier accepts every behavior-preserving control and rejects every invalid shortcut we constructed. The fourteen planned controls on \taskname{HPCG-derived} and \taskname{paged generation} comprise four unchanged-source runs, two legal transformations and eight invalid variants. All unchanged-source runs and both legal transformations pass, and each invalid variant fails the check for the obligation it violates. The controls exercise the named checks and do not estimate a benchmark-wide false-acceptance rate.

\begin{figure}[!htbp]
\centering\includegraphics[width=\linewidth]{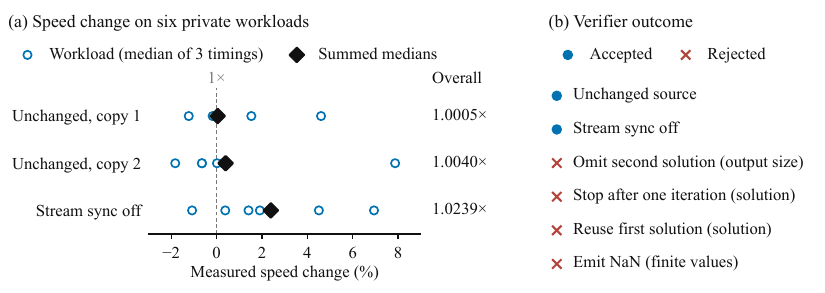}
\caption{\textbf{Unchanged-source controls and verifier challenges on \taskname{HPCG-derived}.} (a) Measured speed change on six private solver workloads for two copies of the unchanged source and for disabling standard-stream synchronization; each circle compares the medians of three timings per side, diamonds and the right column give the ratio of summed medians, and the dashed line marks the theoretical $1\times$. The unchanged-source runs scatter around $1\times$, and disabling stream synchronization passes at $1.024\times$. (b) The verifier accepts both legal programs and rejects all four changes that violate the solution obligations; parentheses name the failed check.}
\label{fig:noop_calibration}
\end{figure}

\paragraph{\taskname{HPCG-derived}: solution obligations.}
The workload solves a sparse system for two right-hand sides. A variant that omits the second solution fails the output-size check, and one that emits NaN fails the finite-value check. Stopping after one iteration or reusing the wrong solution fails the solution checks. The unchanged source and a legal I/O change that disables stream synchronization both pass (Figure~\ref{fig:noop_calibration}).

\paragraph{\taskname{Paged generation}: application checks catch what component checks miss.}
The workload covers full-model computation, checkpoint creation and continuation in a fresh process. Four invalid variants pass the legacy attention-component checks but fail the application checks (Table~\ref{tab:validation}). Lossless checkpoint compression passes both levels.

\begin{table}[!ht]
\caption{\textbf{In \taskname{paged generation}, application checks reject four invalid variants that component checks accept.} All variants run through the original full-workflow verifier. \textcolor{green!45!black}{\ding{51}} denotes acceptance, \textcolor{red!70!black}{\ding{55}} rejection, and --- no speedup for a rejected variant.}
\label{tab:validation}
\centering\small
\begin{tabularx}{\linewidth}{@{}Xccc@{}}
\toprule
Controlled change & Component checks & Application checks & Speedup\\
\midrule
Omit a generation step & \textcolor{green!45!black}{\ding{51}} & \textcolor{red!70!black}{\ding{55}} & ---\\
Advance checkpoint cursor incorrectly & \textcolor{green!45!black}{\ding{51}} & \textcolor{red!70!black}{\ding{55}} & ---\\
Change saved cache metadata & \textcolor{green!45!black}{\ding{51}} & \textcolor{red!70!black}{\ding{55}} & ---\\
Emit nonfinite probabilities & \textcolor{green!45!black}{\ding{51}} & \textcolor{red!70!black}{\ding{55}} & ---\\
Lossless checkpoint compression & \textcolor{green!45!black}{\ding{51}} & \textcolor{green!45!black}{\ding{51}} & 0.894$\times$\\
\bottomrule
\end{tabularx}
\end{table}

\paragraph{Passing is separate from acceleration.}
Both legal transformations pass, yet checkpoint compression measures 0.894$\times$ and disabling stream synchronization 1.024$\times$. This I/O change therefore does not explain the much larger \taskname{HPCG-derived} speedups in the main results.

\paragraph{Measurement protocol.}
All validation controls run on a reserved allocation of four physical cores with the task's memory limit and private workload definitions. Baseline and candidate executions alternate within each workload. The controls add no cache eviction or warmup beyond what the task's evaluator performs. Background services stay active, and we record their observed load.

\section{Evaluation Protocol}\label{sec:eval_protocol}
\rev{This appendix describes the benchmark v1.0 runs of Table~\ref{tab:system_main_v10}; Appendix~\ref{sec:v11_changes} describes v1.1.}

\paragraph{Agents and execution.}
Table~\rev{\ref{tab:system_main_v10}} compares twelve agents, each pairing a model with its harness. GPT-6 Astra runs in Codex CLI and Gemini-3.8-Flash in Gemini CLI, both through a model API gateway. Grok 4.6 runs in \texttt{grok-build}, and the other nine agents run in \texttt{claude-code}. Kimi K3 and GLM-5.3 request \texttt{max} reasoning effort and the other agents \texttt{high}. This provider-specific setting does not equalize inference compute across providers. An open-source agent-evaluation framework orchestrates the runs, which execute in Docker. Tasks allocate four or eight CPU cores and 4--32\,GiB of memory, and each run has an eight-hour limit. Agent runs and verifier evaluations use dual-socket Intel Xeon Platinum 8476C hosts with 48 cores per socket, SMT enabled, two NUMA nodes, AVX-512 and AMX support, and about 2\,TB of memory. During a run, the agent inspects the permitted files, builds candidates and queries the public test. The verifier then rebuilds and scores the final submission. The reward times only the application, measured separately from interaction time.

\paragraph{Run eligibility.}
Runs interrupted by external API failures or infrastructure cancellations are excluded. Agent-caused out-of-memory terminations, including exit code~137, remain eligible and are scored by the verifier like any other final submission.

\paragraph{Reported run.}
Table~\rev{\ref{tab:system_main_v10}} reports one run per agent--task pair, 384 runs in total, and every metric and usage counter in the table comes from that run.

\paragraph{Metrics.}
For the reported run $a^*_{m,t}$ of agent $m$ on task $t$, with correctness $C$ and speedup $s=B_t/T_t$,
\[
\mathrm{Acc}_m=\tfrac{1}{32}\textstyle\sum_t C,\qquad
A_{\tau,m}=\tfrac{1}{32}\textstyle\sum_t C\,\mathbf{1}[s\geq\tau],\qquad
\mathrm{GM}_m=\exp\bigl(\tfrac{1}{32}\textstyle\sum_t\log s^*\bigr),
\]
where $s^*=\max(1,s)$ for correct submissions and $s^*=1$ otherwise. The value $s^*=1$ is an aggregation convention and does not correspond to an observed fallback execution. It keeps unsuccessful runs in the summary without crediting them with a speedup. None of these quantities is conditioned on correctness.

\paragraph{Usage counters.}
A step is one model reply in the archived trajectory. A reply with several tool calls counts once; user messages, system messages and tool returns are excluded. Tokens are the input plus output counts reported by the harness. Input counts include context repeated across calls and cached tokens. Table~\rev{\ref{tab:system_main_v10}} averages these counters over the 32 reported runs of each agent and prints a dash when a harness does not report complete counters. Because provider tokenizers and reasoning-token conventions differ, the counters describe recorded use rather than a normalized compute measure or a verified bill.

\subsection{Model Identifiers}\label{sec:models}
Table~\ref{tab:model_ids} records the API identifiers of the twelve agents in Table~\rev{\ref{tab:system_main_v10}}. We remove provider routing prefixes such as \texttt{xai/} or \texttt{zhipu/}, but show the Gemini identifier as the gateway records it. Some raw Gemini CLI session records retain the client's default alias \texttt{gemini-3.5-flash}; the batch gateway binds this alias to the \texttt{gemini-3.8-flash} backend and records both identities.
\begin{table}[H]
\centering\small
\caption{\textbf{Recorded model identifiers}, in the order of Table~\rev{\ref{tab:system_main_v10}}.}
\label{tab:model_ids}
\begin{tabularx}{\linewidth}{@{}lY@{}}
\toprule Display label & Recorded model identifier\\\midrule
GPT-6 Astra & \texttt{gpt-6-astra}\\
Hy4-Preview & \texttt{hy4-preview}\\
DeepSeek-V4.1-Flash & \texttt{deepseek-v4.1-flash-expires-on-0910}\\
Gemini-3.8-Flash & \texttt{google/gemini-3.8-flash}\\
GLM-5.3 & \texttt{glm-5.3}\\
GLM-5.3-Flash & \texttt{glm-5.3-flash}\\
Grok 4.6 & \texttt{grok-4.6}\\
DeepSeek-V4-Pro & \texttt{deepseek-v4-pro}\\
GLM-5.2 & \texttt{glm-5.2}\\
DeepSeek-V4-Flash & \texttt{deepseek-v4-flash}\\
Kimi K3 & \texttt{kimi-k3}\\
Hy3 & \texttt{hy3}\\
\bottomrule
\end{tabularx}
\end{table}

\section{Timing Calibration and Score Sensitivity}\label{sec:diagnostic_metrics}
For a correct final submission, speedup is $s=B/T$, where the baseline time $B$ and the submission time $T$ use the same task-specific timing scope. Its reward is $r=\max(0,1-1/s)$, and an incorrect submission receives zero (Equation~\ref{eq:runtime_savings}). Each task defines how repeated workload executions are aggregated. The mean reward measures fractional runtime savings over equally weighted tasks, relative to the supplied starting program rather than to an expert implementation. This appendix checks how measurement variation, workload aggregation and alternative score definitions affect these numbers.

\subsection{Timing Calibration}\label{sec:timing_calibration}
\paragraph{Unchanged programs measure within 2\% of 1$\times$.}
An unchanged program has theoretical speedup 1$\times$ and reward zero, but two separate executions can measure on either side of 1$\times$. The \taskname{HPCG-derived} unchanged-source controls measure 1.00049$\times$ and 1.00400$\times$, the \taskname{paged-generation} controls 1.00939$\times$ and 1.00697$\times$, and the \taskname{GROMACS} unchanged-source control 0.99232$\times$. In the \taskname{HPCG-derived} replay of Appendix~\ref{sec:hpcg_replay}, nine unchanged-source evaluations have median 0.9997$\times$ and range 0.9820--1.0165$\times$. Canonical scores keep these raw measurements; Appendix~\ref{sec:reference_sensitivity} examines small positive rewards.

\paragraph{Optimized programs vary more than the baseline.}
In the same replay, one optimized final submission has a 14.54\% coefficient of variation in total runtime across outer rounds. Its median within-shape coefficient of variation across nine process timings is 34.06\%, compared with 3.48\% for the baseline. Noise estimated from the slower original application therefore understates the variation of a short optimized execution. Unchanged-source controls calibrate only the baseline side.

\paragraph{Small rewards are sensitive to the baseline.}
Increasing each baseline time by 1\% while holding candidate times fixed gives 50 previously zero-reward correct submissions a positive reward. This synthetic perturbation uses the timing records of the earlier 842-run snapshot. Of the 625 correct submissions in that snapshot, six lie within $\pm1\%$ of the 1.25$\times$ target. Small rewards thus depend on a reference that is fixed and measured by the verifier. Recomputing rewards from these timing records reproduces all 625 stored values to within $10^{-9}$.

\subsection{Relationships Among Speedup Summaries}
If incorrect and slower submissions fall back to the baseline ($s^*=1$) and $s^*=s$ otherwise, the weighted harmonic mean of $s^*$ equals $1/(1-\bar r)$ under the same weights. The harmonic-mean speedup therefore preserves the Score@1 order and adds no independent criterion. Geometric means and target attainment answer different questions. In Table~\rev{\ref{tab:system_main_v10}}, Hy4-Preview has the highest GM (2.13$\times$), whereas GPT-6 Astra has the highest Score@1, $A_{1.25}$ and $A_2$.

Neighboring benchmarks aggregate speedup differently. SWE-fficiency aggregates expert-relative speedup ratios with a harmonic mean~\citep{swefficiency}. GSO reports success against an expert-relative speed target~\citep{gso}. SWE-Perf measures statistically supported runtime reductions from the original implementation and reports expert patches as references~\citep{sweperf}. Accelerate Bench measures each agent's improvement over the stated starting program. Expert and configuration-tuned implementations are separate calibration references, not upper bounds.

\subsection{Workload Aggregation}\label{sec:workload_aggregation}
Within a task, the ratio of total runtimes is a baseline-weighted harmonic mean of the per-case speedups $s_j=B_j/T_j$:
\[
\frac{\sum_j B_j}{\sum_j T_j}=\left(\sum_j w_j/s_j\right)^{-1},\qquad
w_j=\frac{B_j}{\sum_\ell B_\ell}.
\]
Equal-case harmonic averaging, as in GSO~\citep{gso}, uses different weights and therefore expresses a different workload objective. Neither guarantees that every case improves. For per-case speedups $[0.75,1000]$, the equal-case harmonic mean is about 1.50$\times$ despite a slowdown on the first case. A per-case non-regression requirement needs an explicit gate and a measurement tolerance.

\paragraph{Submissions reaching 2$\times$ improve every measured case.}
In the earlier 842-run snapshot, timing records give complete paired case timings for 421 correct submissions on 23 tasks (1,892 submission--case observations). Their case-time sums reconstruct the scored totals. All 107 submissions reaching 2$\times$ improve every case (Table~\ref{tab:case_aggregation}). Of the 175 reaching 1.25$\times$, eight have a slower case: four in \taskname{HPL}, three in \taskname{NUFFT} and one in \taskname{sparsification}. For only one of the eight do all repeated candidate timings of the slower case exceed the baseline timings. Replacing total-time speedup with an equal-case harmonic mean moves four of the 175 below 1.25$\times$. The analysis covers these 421 submissions; the other 204 correct submissions in that snapshot lack paired case timings.

\begin{table}[!ht]
\centering\small
\caption{\textbf{Workload aggregation changes few target decisions.} Counts use the 421 correct submissions with paired case timings; threshold rows are nested. Slower cases are identified from median case runtimes; the 5\% margin is a sensitivity level, not a noise estimate.}
\label{tab:case_aggregation}
\begin{tabular}{lrrrr}
\toprule
Total speedup & Submissions & Any slower case & Any $>5\%$ slower & Equal-case HM below target\\
\midrule
$>1\times$ & 342 & 102 & 28 & 5\\
$\geq1.25\times$ & 175 & 8 & 5 & 4\\
$\geq2\times$ & 107 & 0 & 0 & 5\\
\bottomrule
\end{tabular}
\end{table}

\subsection{Score Transformations and Reference Choice}\label{sec:reference_sensitivity}
All analyses here use the 384 runs of Table~\rev{\ref{tab:system_main_v10}}, one per agent and task, with tasks weighted equally. Correctness and reward fields are held fixed. Each baseline-fallback speedup is recovered exactly from its reward $r$ as $s^*=1/(1-r)$; this inversion does not reconstruct raw runtimes or the size of slowdowns clipped to zero.

\paragraph{Small gains leave the leading agent unchanged.}
Of the 384 runs, 98 receive zero reward: 59 fail correctness and 39 are correct without acceleration. Another 24 save less than 1\% of baseline runtime. Setting these 24 rewards to zero changes any agent's Score@1 by at most 0.09 percentage points and swaps a single pair, DeepSeek-V4-Flash and Kimi K3, whose Score@1 values differ by 0.07 points. Across savings gates of 0\%, 1\%, 5\%, 10\% and 20\%, each crediting only savings beyond the gate, GPT-6 Astra stays first under both equal-task and equal-area weights. The remaining changes swap agents whose Score@1 values lie within 2 percentage points of each other. We fixed the grid after the outcomes were known, so it is a sensitivity analysis rather than a calibrated noise threshold.

\paragraph{Bounded savings versus log speedup.}
We compare three transformations of the baseline-fallback speedup $s^*$:
\[
 f_{\mathrm{save}}(s^*)=1-1/s^*,\qquad
 f_{\log}(s^*)=\log_2 s^*,\qquad
 f_{\log,K}(s^*)=\min\!\left(1,\frac{\log s^*}{\log K}\right),\quad K>1.
\]
Fractional savings adds progressively less as runtime approaches zero. Log scoring adds the same increment for each doubling, and the capped form stops rewarding gains beyond $K$.

\begin{table}[!ht]
\centering\small
\caption{\textbf{Score transformations change the emphasis on large gains.} Example scores assume correctness. The final column is the share of total additive score from HPCG and paged generation over the 384 runs of Table~\rev{\ref{tab:system_main_v10}}, weighting tasks and agents equally. The caps $K$ are sensitivity choices.}
\label{tab:metric_transforms}
\begin{tabular*}{\linewidth}{@{\extracolsep{\fill}}lrrrr@{}}
\toprule
& \multicolumn{3}{c}{Score at a given speedup} & Contribution\\
\cmidrule(lr){2-4}
Scoring rule & $2\times$ & $4\times$ & $8\times$ & of the two tasks\\
\midrule
Runtime savings & 0.500 & 0.750 & 0.875 & 22.1\% \\
Log speedup & 1.000 & 2.000 & 3.000 & 38.4\% \\
Capped log, $K=2$ & 1.000 & 1.000 & 1.000 & 15.7\% \\
Capped log, $K=4$ & 0.500 & 1.000 & 1.000 & 22.7\% \\
Capped log, $K=8$ & 0.333 & 0.667 & 1.000 & 29.4\% \\
Capped log, $K=16$ & 0.250 & 0.500 & 0.750 & 34.4\% \\
Capped log, $K=32$ & 0.200 & 0.400 & 0.600 & 36.9\% \\
\bottomrule
\end{tabular*}
\end{table}

Six of the seven rules keep GPT-6 Astra first. Uncapped log speedup ranks Hy4-Preview first, consistent with its higher GM in Table~\rev{\ref{tab:system_main_v10}}. Every other change reorders agents whose Score@1 values lie within 2 percentage points. The rules differ mainly in the weight they give to large gains. With equal task and agent weights, \taskname{HPCG} and \taskname{paged generation} contribute 22.1\% of total bounded savings, and uncapped log scoring raises their share to 38.4\%; under capped log scores, the share rises from 15.7\% at $K=2$ to 36.9\% at $K=32$ (Table~\ref{tab:metric_transforms}). These shares refer to additive scores, not fractions of GM speedup. Bounded savings caps any single task's contribution to the equal-task mean at $1/32$.

\paragraph{Replacing the baseline with the best observed speedup.}
Let $U_t$ be the largest correct speedup among the 384 reported runs on task $t$. $U_t$ is a pooled observed reference, not an expert or optimal runtime. All 32 tasks have $U_t>1$. With the same task weights, we compare
\begin{align*}
 H_B&=\langle 1/s^*\rangle^{-1}, &
 H_{\mathrm{pool}}&=\langle U_t/s^*\rangle^{-1},\\
 P_{\mathrm{save}}&=\left\langle\frac{1-1/s^*}{1-1/U_t}\right\rangle, &
 P_{\log}&=\left\langle\frac{\log s^*}{\log U_t}\right\rangle.
\end{align*}
$H_B$ is the harmonic speedup over the supplied baseline, and $H_{\mathrm{pool}}$ is the harmonic ratio to the pooled reference. Unsuccessful runs keep $s^*=1$ and hence fallback performance $1/U_t$. The progress scores map the baseline to zero and the pooled reference to one.

\begin{table}[!ht]
\centering\small
\caption{\textbf{Reference choice changes the meaning of the score.} All columns use the 384 runs of Table~\rev{\ref{tab:system_main_v10}}. $H_B$ measures harmonic speedup from the supplied baseline; $H_{\mathrm{pool}}$ measures the harmonic ratio to each task's best observed speedup. $P_{\mathrm{save}}$ and $P_{\log}$ normalize progress to that observed reference. The pooled reference is the best speedup among these agents, not an expert baseline.}
\label{tab:reference_choices}
\begin{tabular*}{\linewidth}{@{\extracolsep{\fill}}lrrrrr@{}}
\toprule
Agent & Score@1 (\%) & $H_B$ ($\times$) & $H_{\mathrm{pool}}$ & $P_{\mathrm{save}}$ (\%) & $P_{\log}$ (\%)\\
\midrule
GPT-6 Astra & 40.3 & 1.675 & 0.790 & 71.3 & 67.3 \\
Hy4-Preview & 36.5 & 1.574 & 0.806 & 62.9 & 60.4 \\
DeepSeek-V4.1-Flash & 29.9 & 1.427 & 0.580 & 50.2 & 46.2 \\
Gemini-3.8-Flash & 29.9 & 1.427 & 0.583 & 53.5 & 47.4 \\
GLM-5.3 & 28.1 & 1.391 & 0.570 & 48.0 & 44.6 \\
GLM-5.3-Flash & 27.3 & 1.376 & 0.560 & 53.4 & 47.2 \\
Grok 4.6 & 26.2 & 1.356 & 0.578 & 44.9 & 39.4 \\
DeepSeek-V4-Pro & 20.6 & 1.260 & 0.473 & 32.0 & 26.2 \\
GLM-5.2 & 19.9 & 1.249 & 0.484 & 31.2 & 25.4 \\
DeepSeek-V4-Flash & 19.1 & 1.236 & 0.453 & 31.1 & 25.6 \\
Kimi K3 & 19.0 & 1.235 & 0.211 & 32.9 & 27.6 \\
Hy3 & 9.8 & 1.108 & 0.394 & 15.5 & 12.3 \\
\bottomrule
\end{tabular*}
\end{table}

The progress scores keep GPT-6 Astra and Hy4-Preview in the top two places and Hy3 last, but reorder the agents between them (Table~\ref{tab:reference_choices}). The pooled reference ranges from 1.002$\times$ on \taskname{Fluidity} to 89.46$\times$ on \taskname{HPCG}, so normalizing to it maps a small and a large measured gain to the same endpoint. On \taskname{ICON}, the savings denominator $1-1/U_t$ is about 0.021 and amplifies small timing changes. Tasks with large $U_t$ dominate $H_{\mathrm{pool}}$. A run left at baseline on \taskname{HPCG} contributes $1/89.46$, which moves Kimi K3, whose reported \taskname{HPCG} run earns zero reward, to last place. A common per-task reference cancels from geometric means, $\mathrm{GM}(s^*/U_t)=\mathrm{GM}(s^*)/\mathrm{GM}(U_t)$, so it leaves the order on a shared panel unchanged. Harmonic and log normalizations instead reweight tasks.

\paragraph{An independently written reference.}\label{sec:hpcg_reference}
An expert implementation with measured speedup $e_t$ could replace $U_t$, for example as progress $\log s^*/\log e_t$, provided it passes the same checks under the same resource, workload and timing rules and is fixed independently of the agents compared. We built one such reference for \taskname{HPCG-derived}. The task's operator is $A=27I-S_x\otimes S_y\otimes S_z$, where each $S$ has ones on the main and adjacent diagonals, and the sine basis diagonalizes each factor. Our implementation applies dense orthonormal sine transforms along the three axes and divides by the eigenvalues. It uses no external FFT library and modifies only \texttt{src/application\_main.cpp}. We wrote it with knowledge of the task and of earlier trajectories, so it is a separate implementation rather than a blind discovery or an expert optimum.

We evaluated three artifacts: the pristine source, this reference and one agent's final submission (DeepSeek-V4-Flash, \texttt{45WWLii}). Each receives three clean builds and full private evaluations on four physical cores within one NUMA node, with SMT siblings excluded, 6\,GiB of memory and no network. The verifier image, tests and six workload shapes are unchanged; one new right-hand-side seed is used throughout, and artifact order is counterbalanced across rounds. All nine evaluations pass. The reference reaches 21.18--22.22$\times$ and the agent's submission 25.86--26.33$\times$. Paired by round, the agent's program is 1.19--1.25$\times$ faster than ours in every round (Table~\ref{tab:hpcg_reference}). The three-round ranges describe these measurements, not population intervals. This replay does not change official scores.

\begin{table}[ht]
\centering\small
\caption{\textbf{The agent's \taskname{HPCG-derived} solver outperforms an independently written direct solver.} The first three rows are speedups over the supplied baseline; the last row divides the reference's runtime by the agent's within each round. All nine evaluations pass.}
\label{tab:hpcg_reference}
\begin{tabular}{lrrr}
\toprule
Comparison & Round 1 & Round 2 & Round 3\\
\midrule
Rebuilt pristine source & 1.004$\times$ & 0.996$\times$ & 1.001$\times$\\
Independent direct solver & 21.180$\times$ & 22.216$\times$ & 21.261$\times$\\
Agent final submission & 26.325$\times$ & 25.862$\times$ & 26.209$\times$\\
Independent time / agent time & 1.245$\times$ & 1.189$\times$ & 1.200$\times$\\
\bottomrule
\end{tabular}
\end{table}

\section{Per-Task Results}\label{sec:taskresults}
Large speedups concentrate on a subset of tasks. Table~\ref{tab:task_results} lists the per-task reward of every run in Table~\rev{\ref{tab:system_main_v10}} \rev{(benchmark v1.0)}. At least one agent reaches $2\times$ on 15 of the 32 tasks, but all twelve do so only on \taskname{Paged LM generation}. On seven tasks (\taskname{Fluidity}, \taskname{ICON}, \taskname{miniVite}, \taskname{NEMO}, \taskname{Quantum ESPRESSO}, \taskname{QuEST} and \taskname{VPIC}), no agent reaches $1.25\times$. Correctness failures are most frequent on \taskname{Fluidity}, where seven of the twelve runs fail.

\begin{table}[!htbp]
\centering\footnotesize\setlength{\tabcolsep}{1.6pt}
\caption{\textbf{Per-task reward of the runs in Table~\rev{\ref{tab:system_main_v10}}.} Entries are runtime savings $r_t$ (\%); \texttimes{} marks a run that fails correctness checks, and 0 a correct run without acceleration. Bold entries reach at least $2\times$ speedup ($r_t\geq50$); entries just below the 20 and 50 thresholds show decimals. Agents are ordered by Score@1, the column mean.}
\label{tab:task_results}
\begin{tabular*}{\linewidth}{@{\extracolsep{\fill}}lrrrrrrrrrrrr@{}}
\toprule
\rule{0pt}{5.2em}Task & \multicolumn{1}{c}{\rlap{\rotatebox{60}{GPT-6 Astra}}} & \multicolumn{1}{c}{\rlap{\rotatebox{60}{Hy4-Preview}}} & \multicolumn{1}{c}{\rlap{\rotatebox{60}{DeepSeek-V4.1-Flash}}} & \multicolumn{1}{c}{\rlap{\rotatebox{60}{Gemini-3.8-Flash}}} & \multicolumn{1}{c}{\rlap{\rotatebox{60}{GLM-5.3}}} & \multicolumn{1}{c}{\rlap{\rotatebox{60}{GLM-5.3-Flash}}} & \multicolumn{1}{c}{\rlap{\rotatebox{60}{Grok 4.6}}} & \multicolumn{1}{c}{\rlap{\rotatebox{60}{DeepSeek-V4-Pro}}} & \multicolumn{1}{c}{\rlap{\rotatebox{60}{GLM-5.2}}} & \multicolumn{1}{c}{\rlap{\rotatebox{60}{DeepSeek-V4-Flash}}} & \multicolumn{1}{c}{\rlap{\rotatebox{60}{Kimi K3}}} & \multicolumn{1}{c}{\rlap{\rotatebox{60}{Hy3}}}\\
\midrule
GROMACS & \textbf{65} & \textbf{63} & 0 & \textbf{58} & \textbf{52} & \textbf{53} & \textbf{53} & \textbf{52} & \textbf{53} & \textbf{53} & \textbf{53} & 0 \\
LAMMPS & \textbf{54} & 47 & \textbf{56} & 1 & 3 & 7 & 11 & 0 & 1 & 0 & 7 & 3 \\
GPAW & 31 & 5 & 4 & 23 & \texttimes & \texttimes & \texttimes & 12 & \texttimes & 12 & \texttimes & 0 \\
Quantum ESPRESSO & 9 & \texttimes & 0 & 4 & 12 & 4 & \texttimes & 1 & 1 & 1 & 0 & \texttimes \\
SIESTA & \textbf{61} & \textbf{55} & \texttimes & 39 & 2 & 49.8 & \texttimes & \texttimes & \texttimes & \textbf{51} & 49.7 & \texttimes \\
\addlinespace[2pt]
Fluidity & 0 & 0 & \texttimes & \texttimes & \texttimes & 0 & \texttimes & \texttimes & 0 & \texttimes & 0 & \texttimes \\
ICON & 0 & 2 & \texttimes & 0 & 0 & 2 & \texttimes & \texttimes & 0 & 0 & 0 & 0 \\
MPAS-A & 23 & 25 & 4 & 24 & 19 & 14 & 1 & \texttimes & 0 & 2 & 24 & \texttimes \\
NEMO & 0 & 3 & 1 & \texttimes & \texttimes & 3 & 2 & \texttimes & 3 & 1 & 1 & 3 \\
OpenFOAM & 23 & 12 & 7 & 12 & 0 & 2 & 2 & 25 & 0 & 13 & \texttimes & 0 \\
Palabos & \textbf{88} & \textbf{91} & 0 & \textbf{62} & \texttimes & 16 & \textbf{72} & 23 & \texttimes & 1 & 11 & 0 \\
SU2 & \textbf{56} & \textbf{64} & \textbf{61} & 45 & 44 & 29 & 25 & \textbf{54} & 32 & 13 & 31 & 8 \\
\addlinespace[2pt]
OpenMC & 47 & 30 & 21 & 26 & 37 & 3 & 4 & 3 & 9 & 17 & 15 & 3 \\
ShengBTE & \textbf{54} & \texttimes & \textbf{55} & \texttimes & 43 & 38 & 37 & 36 & 36 & 35 & 15 & \texttimes \\
VPIC & 0 & 0 & \texttimes & 5 & \texttimes & 3 & 2 & 0 & \texttimes & \texttimes & \texttimes & 0 \\
PRESTO & 27 & 0 & 0 & 1 & \texttimes & \texttimes & 0 & 0 & \texttimes & 2 & \texttimes & \texttimes \\
\addlinespace[2pt]
ABySS & \texttimes & 42 & 49 & \texttimes & 49.9 & 45 & 43 & 36 & 10 & 17 & 37 & 9 \\
MHM2 & 37 & 2 & \texttimes & 33 & \texttimes & 27 & 2 & 0 & \texttimes & 0 & 0 & 3 \\
MrBayes & 29 & 32 & 23 & 12 & 11 & 19 & 9 & 14 & 12 & 7 & 17 & 8 \\
wtdbg2 & \textbf{58} & \textbf{58} & 39 & 16 & 13 & 0 & 34 & 22 & 0 & 24 & 0 & 0 \\
\addlinespace[2pt]
HPCG-derived & \textbf{97} & \textbf{99} & \textbf{98} & \textbf{92} & \textbf{99} & \textbf{97} & \textbf{99} & \textbf{97} & \textbf{98} & \textbf{98} & \texttimes & \textbf{97} \\
HPL-derived & 45 & 34 & 32 & 34 & 47 & 34 & \textbf{68} & 32 & 48 & 22 & 36 & 7 \\
NUFFT imaging & \textbf{80} & \textbf{52} & \textbf{75} & \textbf{76} & \texttimes & 44 & 47 & 38 & 32 & 33 & \textbf{54} & 33 \\
Tucker reconstruction & \textbf{68} & \textbf{82} & \textbf{63} & \textbf{74} & \textbf{70} & \textbf{58} & 41 & 36 & 33 & 30 & \texttimes & 31 \\
Graph sparsification & \textbf{63} & \textbf{79} & \textbf{79} & 16 & \textbf{79} & \textbf{78} & \textbf{58} & 32 & \textbf{61} & 34 & 32 & 14 \\
miniVite & 10 & 18 & 19 & 8 & 14 & 12 & 19.95 & 4 & 7 & 3 & 6 & 2 \\
\addlinespace[2pt]
QuEST & 12 & 10 & 11 & 9 & 12 & 0 & 16 & 0 & 7 & 9 & 0 & 2 \\
QiboTN & \texttimes & 0 & 29 & \texttimes & 31 & \texttimes & 34 & 2 & 0 & 1 & 0 & 0 \\
\addlinespace[2pt]
Paged LM generation & \textbf{94} & \textbf{94} & \textbf{94} & \textbf{93} & \textbf{94} & \textbf{91} & \textbf{93} & \textbf{93} & \textbf{91} & \textbf{90} & \textbf{85} & \textbf{87} \\
ThunderSVM & \textbf{77} & \textbf{85} & \textbf{53} & \textbf{70} & \textbf{83} & \textbf{65} & 8 & 48 & 43 & 40 & \textbf{52} & \texttimes \\
\addlinespace[2pt]
DataLife & \textbf{78} & \textbf{84} & \textbf{83} & \textbf{87} & \textbf{83} & \textbf{79} & \textbf{59} & \texttimes & \textbf{62} & 1 & \textbf{82} & 1 \\
IO500-derived & 4 & 2 & 3 & 37 & 0 & 1 & 1 & 0 & \texttimes & \texttimes & 0 & 1 \\
\midrule
Score@1 & 40.3 & 36.5 & 29.9 & 29.9 & 28.1 & 27.3 & 26.2 & 20.6 & 19.9 & 19.1 & 19.0 & 9.8 \\
\bottomrule
\end{tabular*}
\end{table}

\section{Annotation Procedures}\label{sec:annotation_methods}
Section~\ref{sec:analysis} uses three kinds of labels (Table~\ref{tab:annotation_sources}). Technique labels describe what a run's final implementation changes relative to the supplied source. Full-trajectory annotations describe each tool call in a run and the relationships among the optimization choices its trajectory records. Measurement-command groups come from fixed lexical rules (Appendix~\ref{sec:measurement_groups}).

\begin{table}[!htb]
\centering
\caption{Label sources used in Section~\ref{sec:analysis}. All three cover the same 480 runs.}
\label{tab:annotation_sources}
\small
\begin{tabular}{@{}p{0.15\linewidth}p{0.22\linewidth}p{0.19\linewidth}p{0.2\linewidth}p{0.11\linewidth}@{}}
\toprule
Labels & Annotator & Unit & Input & Used in \\
\midrule
Technique & GPT-5.6 Luna, medium effort, Codex CLI & Final implementation of a run & Supplied and candidate source, build and launch files, trajectory excerpts & \S\ref{sec:rq1} \\
Full trajectory & DeepSeek-V4.1-Flash, high effort, Responses API & Tool call; relationship record & Complete trajectory text & \S\ref{sec:rq2}, \S\ref{sec:mechanism_controls} \\
Measurement command & Lexical rules & Run & Shell-command arguments & \S\ref{sec:workflow_measurement} \\
\bottomrule
\end{tabular}
\end{table}

\subsection{Technique Labels}\label{sec:technique_annotation}
GPT-5.6 Luna, run through Codex CLI at medium reasoning effort, reviews one task at a time. Each review covers every run of that task in a labeling panel that includes all 480 runs analyzed here. The reviewer reads the supplied source, the retained candidate source and launch configuration, and the trajectory. Each source excerpt cited as direct evidence keeps its file path and line range.

Each review has two phases. First, the reviewer writes a task map: what the application computes, what must remain correct, and what the supplied implementation executes. It then defines task-local algorithm routes and mechanism tests and labels every run. These labels are fixed before the reviewer sees the scored outcomes. Second, the reviewer checks each outcome record against the saved labels and records any conflict as a separate proposed correction.

For each run, the review records a class (optimization, original scheme, or unresolved), mechanism tags from a shared vocabulary of fifteen optimization mechanisms and one correctness-repair tag, implementation status, confidence, source-access level, and evidence locators. A change counts only if it is active in the final execution path, so unwired, compiled-out, or reverted code does not qualify. A correctness or build repair alone is not an optimization. Changing a thread count is not labeled as a new parallel algorithm. In the 480-run panel, 464 labels rest on directly read final source and 15 on source excerpts bound to the final state; 438 have high confidence.

An independent source audit then mapped the fifteen mechanism tags into six categories and checked tags against recorded behavior. It corrected tags in nine runs and withheld three labels whose active execution path could not be confirmed. With these three labels withheld and one left unresolved by the review, 476 runs are classified.

The six categories are computation and reuse (A1--A7), parallelism and vectorization (T1--T3), runtime and numerical configuration (R1--R2), library/backend selection (L1), compiler/build tuning (B1), and I/O (I1). Of the 476 classified runs, 361 have at least one retained optimization label; the other 115 keep the original scheme, possibly with repairs or other edits. The four unclassified runs remain in the performance and command analyses and are excluded from technique-label contrasts. A run may carry multiple labels.

\paragraph{Boundaries between related labels.}
Numerical reformulation (A2) changes the mathematical formulation or equations, whereas numerical parameter choice (R2) changes exposed tolerances, discretization, preconditioner, or solver settings. Memoization/precomputation (A4) retains computed values or hoists invariant work; workspace reuse (A6) retains allocated storage across iterations. The task codebooks state these distinctions explicitly, for example in the \taskname{OpenFOAM} definitions. CPU parallelism (T1) records changes to active parallel execution, while runtime resource tuning (R1) records resource settings. This boundary varies across codebooks. \taskname{OpenFOAM} separates explicit solver threading from resource settings, whereas \taskname{paged LM generation} places active thread-pool and BLAS/OpenMP controls under T1. The reported counts keep these task-local assignments.

\subsection{Full-Trajectory Annotation}\label{sec:trajectory_annotation_method}
DeepSeek-V4.1-Flash annotates the complete trajectories of all 480 runs at high reasoning effort, through a Responses API gateway with a strict JSON output schema and up to 131,072 output tokens per request. The model receives the full trajectory text, including every step, message, reasoning record, tool argument, and tool return, but not final scores or earlier annotations.

Trajectories are split without dropping text. Each field is cut into segments of at most 10,000 characters, and each segment keeps its position in the original field. The segments are grouped into 11,820 core chunks of about 144,000 characters, keeping whole steps together where possible. Each request adds 24,000 characters of context per side. A parser assigns every tool call an action ID and every message without a tool call a statement ID; the annotator labels each ID in its core chunk exactly once.

For each action, the annotator records its facets, such as source edit, build configuration, correctness test, performance measurement, or profiling; a decision label (investigate, propose, implement, refine, repair, revert, retain/select, test, submit, continue, no new decision, or unclear); the basis for that decision; the visible effect of the action; and a selection label. Reasons, cited feedback, and results point to earlier text spans. An action is labeled successful when the requested operation succeeded; correctness and speed are taken from returned tests. A revert is labeled only when visible in an action. A relationship record connects two distinct optimization choices, or an optimization and an execution-path condition that enables, blocks, or bypasses it. It carries one of six types and a basis in visible source, returned feedback, or an agent statement. Performance hypotheses are recorded as agent claims.

Before the full run, we validated the protocol on 24 inputs: 22 real chunks, a task-description negative control, and a repeated chunk. On the repeated chunk, decision labels agree for 19 of 21 actions and selection labels for all 21. All 11,820 chunk outputs pass checks for schema validity, exact action coverage, valid evidence IDs, source-role constraints, and chronology.

Chunk outputs are assembled into one evidence graph per trajectory, giving 203,472 tool actions, 417 agent-statement records, and 23,845 relationship records. Exact duplicate records are merged, and simultaneous tool calls share their original agent turn. Every cited text span is checked against its source. Targeted source reviews check the distinctions the analysis relies on: plans versus executed operations, agent hypotheses versus returned observations, temporary working copies versus the submitted tree, and tool success versus application correctness. Recomputing from the saved evidence graphs reproduces every count and association in Sections~\ref{sec:rq2} and~\ref{sec:mechanism_controls}.

Decision labels are correct for 77 of 80 calls (96\%; 95\% Wilson interval 90--99\%), and the test, performance-measurement, profiling, and source-edit facets for 73 (91\%; 83--96\%). Cited reasons explain the labeled call in 67 of 71 applicable cases (94\%), and cited feedback is output the call responds to in 63 of 66 (95\%). Errors concentrate in the measurement stratum (16 of 20 facets correct); most are calls that prepare profiling, such as adding \texttt{-pg}, but are labeled as profiling. Relationship records are less reliable, so Section~\ref{sec:mechanism_controls} uses them only to locate evidence for review (Appendix~\ref{sec:dependency_review}).

\subsection{From Labels to Analysis Variables}\label{sec:annotation_models}
Each analysis converts labels into run-level variables before comparing outcomes. Technique contrasts compare labeled and unlabeled runs of the same model and task, average within tasks, and weight eligible tasks equally, with task-bootstrap intervals (Appendix~\ref{sec:technique_contrasts}). Measurement-command groups enter a regression with additive task and model effects (Appendix~\ref{sec:measurement_groups}). Trajectory-feature models add a fixed effect per model--task cell (Appendix~\ref{sec:trajectory_outcomes}). All three report task-clustered 95\% intervals.

\section{Technique and Workflow Analysis}\label{sec:optimization_panel}
This appendix gives the sample, definitions and full tables behind Sections~\ref{sec:rq1} and~\ref{sec:rq2}, in the order of their observations. Appendix~\ref{sec:annotation_methods} describes how the labels are assigned.

\subsection{Analysis Sample}\label{sec:optimization_sample}
\begin{figure}[H]
\centering
\includegraphics[width=5.5in]{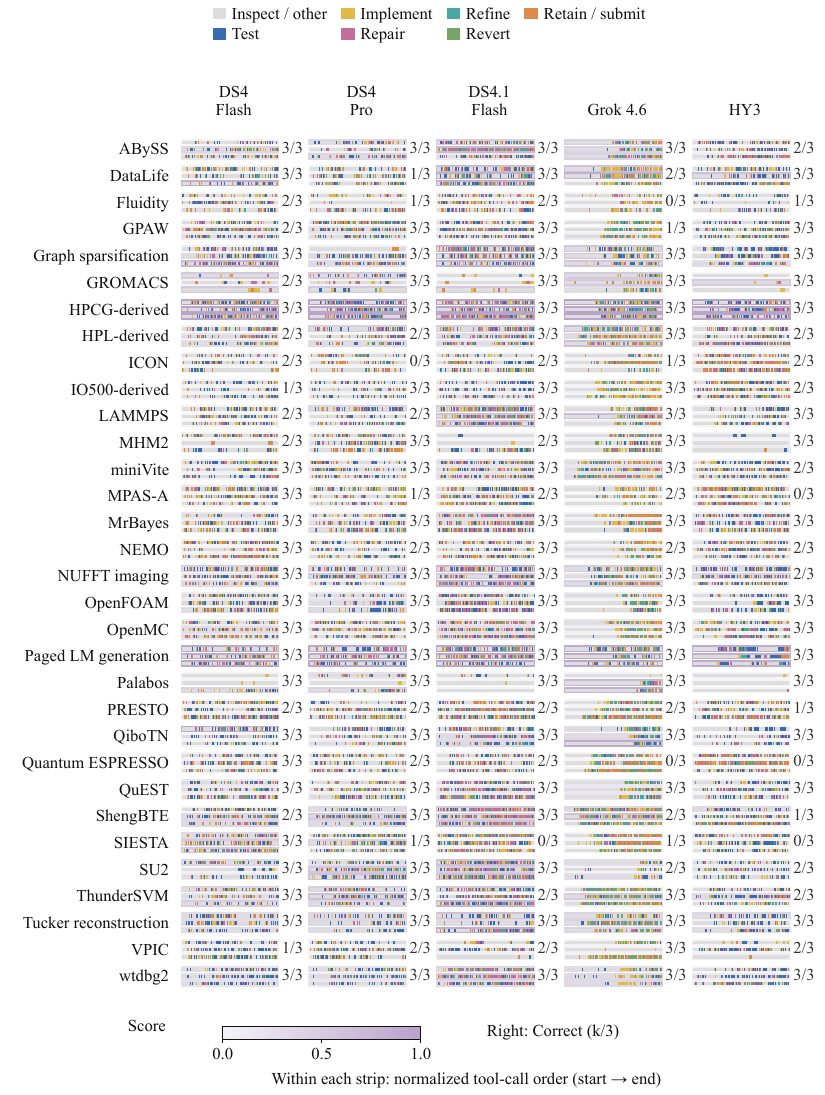}
\caption{\textbf{Action sequences and outcomes across all 480 analysis runs.} Each task--agent cell contains three runs, ordered by trial ID. Foreground colors show tool-action annotations; the pale background of each strip shows its final, correctness-gated runtime-saving score on a common 0--1 scale. A score of zero can also indicate a correct program without a runtime saving. Counts at right report correct runs out of three. Horizontal position shows recorded tool-call order, normalized separately for each run, not elapsed time; calls in the same recorded step may overlap. The sequences contain 203,472 tool actions; 417 statement-only annotations are not plotted.}
\label{fig:trajectory_atlas}
\end{figure}

Each of the five agents of Section~\ref{sec:analysis_data} runs three times on all 32 tasks, giving 480 runs that represent every model--task pair equally and have complete annotation and command records. Of the final submissions, 396 are correct and 84 incorrect; 175 of the correct ones reach $1.25\times$. Final performance uses the scored outcomes, and workflow analyses use the public-test responses and timings recorded during each run. Figure~\ref{fig:trajectory_atlas} shows every annotated action sequence alongside its outcome.

\subsection{Technique Counts and Outcome Contrasts}\label{sec:technique_counts}\label{sec:technique_contrasts}
Tables~\ref{tab:optimization_outcomes} and~\ref{tab:optimization_models} give technique counts by final outcome and by model for the 476 classified runs.

\begin{table}[!htb]
\centering
\caption{Optimization labels and final outcomes in the 480-run panel. Categories overlap; 476 runs have classified technique labels. Speedup requires final correctness.}
\label{tab:optimization_outcomes}
\small
\begin{tabular}{lrrrr}
\toprule
Technique & Runs & Tasks & Correct & $\geq1.25\times$ \\
\midrule
Computation / reuse & 225 & 29 & 201 (89.3\%) & 108 (48.0\%) \\
Parallel / vectorized & 132 & 22 & 116 (87.9\%) & 83 (62.9\%) \\
Runtime / numerical & 64 & 17 & 57 (89.1\%) & 31 (48.4\%) \\
Library / backend & 46 & 11 & 39 (84.8\%) & 26 (56.5\%) \\
Compiler / build & 40 & 14 & 32 (80.0\%) & 10 (25.0\%) \\
I/O / serialization & 28 & 8 & 24 (85.7\%) & 11 (39.3\%) \\
\bottomrule
\end{tabular}
\end{table}

\begin{table}[!htb]
\centering
\caption{Eligible source labels and technique counts by model. C: computation/reuse; P: parallelism/vectorization; R: runtime/numerical configuration; L: library/backend; B: compiler/build; I: I/O. Each model has 96 scored runs; categories overlap.}
\label{tab:optimization_models}
\small
\begin{tabular}{lrrrrrrr}
\toprule
Model & Labeled & C & P & R & L & B & I \\
\midrule
DeepSeek-V4.1-Flash & 96 & 45 & 34 & 8 & 7 & 5 & 7 \\
Grok 4.6 & 95 & 67 & 36 & 6 & 7 & 3 & 13 \\
DeepSeek-V4-Pro & 93 & 33 & 26 & 17 & 8 & 13 & 3 \\
DeepSeek-V4-Flash & 96 & 46 & 22 & 16 & 11 & 8 & 2 \\
Hy3 & 96 & 34 & 14 & 17 & 13 & 11 & 3 \\
\bottomrule
\end{tabular}
\end{table}

For each category, we compare mean final reward between runs with and without the label, within each model--task group that contains both. Group contrasts are averaged within each task, and eligible tasks receive equal weight; incorrect runs keep their scored reward. Intervals resample task means 10,000 times with seed 20260922. Table~\ref{tab:optimization_contrasts} reports each category's group and task counts, estimate and interval.

\begin{table}[!htb]
\centering
\caption{Exploratory reward contrasts within model--task groups, averaged over eligible tasks. Intervals use 10,000 task-bootstrap resamples and are pointwise, not simultaneous.}
\label{tab:optimization_contrasts}
\small
\begin{tabular}{lrrr}
\toprule
Technique & Groups & Tasks & Difference (pp) [95\% CI] \\
\midrule
Computation / reuse & 53 & 27 & +8.0 [1.4, 15.9] \\
Parallel / vectorized & 46 & 19 & +8.9 [1.9, 17.0] \\
Runtime / numerical & 35 & 17 & +0.6 [-6.3, 8.9] \\
Library / backend & 15 & 10 & -4.9 [-28.0, 14.4] \\
Compiler / build & 25 & 13 & -0.7 [-6.4, 6.4] \\
I/O / serialization & 9 & 7 & +1.7 [-8.1, 12.4] \\
\bottomrule
\end{tabular}
\end{table}

\subsection{Five-Agent Process Comparison}\label{sec:trajectory_systems}
Figure~\ref{fig:trajectory_overview} uses all 203,472 tool-action annotations. Within each trajectory, we compute category fractions and split the tool-call sequence into three consecutive, nearly equal parts; the test-label fraction of a part uses that part's calls as the denominator. Each agent's value weights its 96 trajectories equally. Retention and submission form one displayed category, and investigation, continuation, proposals, unclear actions and other labels form the inspection/other category.

A multi-tool turn is a tool-bearing step with more than one call. Each run's fraction divides these turns by all its tool-bearing steps. Averaged over runs, the fraction is 57.1\% for Grok 4.6, compared with 10.0\% for DeepSeek-V4-Flash, 8.8\% for DeepSeek-V4-Pro, 9.4\% for DeepSeek-V4.1-Flash and 1.6\% for Hy3. Intervals use 10,000 task-bootstrap draws with seed 20260924.

The DeepSeek and Hy3 records identify Claude Code 2.1.265 (one DeepSeek-V4-Pro record lacks harness metadata), and all Grok records identify Grok Build 1.0.24. Grok comparisons therefore vary both model and harness. Within Claude Code, DeepSeek-V4.1-Flash reaches $1.25\times$ in 46 of 96 runs and DeepSeek-V4-Flash in 35 ($+11.5$ percentage points; 95\% interval $[-2.1,24.0]$). The outcome panel counts all 96 runs per agent, including incorrect ones.

\subsection{Timing and Profiling Groups}\label{sec:measurement_groups}
Lexical rules scan executable shell-command arguments over each full trajectory. Runs with a profiling-related match form P2 and runs with only timing matches form P1; P0 runs have no matched command but may measure performance in other ways. A matched command need not complete or address the scored workload. Table~\ref{tab:optimization_diagnosis} covers all 480 runs. Raw reward intervals resample the 32 tasks 10,000 times. Adjusted contrasts regress reward on group indicators with additive task and model effects, using a CR1 task-cluster covariance and $t_{31}$ intervals.

Adjusted mean reward differences are $+4.4$ percentage points for P1--P0 (95\% interval $[-4.0,12.8]$), $+8.1$ for P2--P0 ($[-2.9,19.0]$) and $+3.7$ for P2--P1 ($[-0.6,8.0]$). Adjustment reduces the P2--P0 difference from 17.9 points to 8.1, so more than half of the raw gap reflects which tasks and models profile.

\begin{table}[!htb]
\centering
\caption{Command-based measurement groups across all 480 trajectories. P0, P1, and P2 are observed tool patterns, not ordered levels of diagnosis quality.}
\label{tab:optimization_diagnosis}
\small
\begin{tabular}{lrrrr}
\toprule
Group & Runs & Correct & $\geq1.25\times$ & Mean reward \\
\midrule
P0 & 61 & 48 & 7 & 8.1\% \\
P1 & 251 & 197 & 93 & 22.2\% \\
P2 & 168 & 151 & 75 & 26.0\% \\
\bottomrule
\end{tabular}
\end{table}

\subsection{Comparable Version Histories}\label{sec:version_histories}
A version history is the sequence of correct candidates in one run that share a timing protocol. Timing comparisons require an original correct response, a finite positive runtime in the same field, source-confirmed contents and matching protocol fields: profiles, case counts, grid shapes, graph sizes, right-hand-side counts and prompt lengths where reported. Two candidates count as the same program when their files and links are identical. In each run, we keep the largest comparable group, breaking ties by first observation. Thirty-one histories contain at least two distinct programs: ten each from DeepSeek-V4-Pro and DeepSeek-V4-Flash, seven from DeepSeek-V4.1-Flash and four from Grok 4.6. Hy3 has none.

\paragraph{Endpoints.}
Each program's runtime is the median of its observed runtimes. The first and last programs follow request order, and the best program has the lowest median. We report $T_{\mathrm{first}}/T_{\mathrm{best}}$ and $T_{\mathrm{first}}/T_{\mathrm{last}}$, and the last/best gap $T_{\mathrm{last}}/T_{\mathrm{best}}$. At threshold $\delta$, a first-to-last gain requires $T_{\mathrm{first}}/T_{\mathrm{last}}>1+\delta$, a regression $T_{\mathrm{last}}/T_{\mathrm{first}}>1+\delta$, and a last/best gap $T_{\mathrm{last}}/T_{\mathrm{best}}>1+\delta$. The first program is the first correct candidate, not the task baseline. The last correct public test can be followed by failed requests or untested edits. The final submission's source files match the last program in 25 histories, ten of them with a last/best gap above 5\%.

\paragraph{Consecutive changes.}
The 70 changed-program pairs come from 26 of these histories. Both programs in a pair must be adjacent in the original sequence, source-confirmed, correct and protocol-comparable; resubmitted programs are excluded. Each ratio compares the two observed runtimes. A revision is faster when $T_{\mathrm{previous}}/T_{\mathrm{current}}>1$, and it sets a new best when it beats every earlier comparable correct observation.

\begin{table}[!htbp]
\centering
\caption{Public iteration outcomes within the 480-run panel. Rates first average revisions within runs, then runs within applications; 95\% intervals resample the seven represented applications.}
\label{tab:optimization_workflow}
\small
\begin{tabular}{lrr}
\toprule
Observation & Pooled count & App.-weighted \% [95\% CI] \\
\midrule
First-to-last speedup $>1.05\times$ & 19/31 & 65.2 [40.5, 86.6] \\
Last is $>5\%$ slower than best & 11/31 & 31.3 [9.5, 58.1] \\
Last is the best tested version & 14/31 & 51.3 [29.7, 73.2] \\
Revision improves over predecessor & 36/70 & 62.9 [45.8, 81.2] \\
Revision sets a new observed best & 26/70 & 58.5 [40.0, 78.4] \\
\bottomrule
\end{tabular}
\end{table}

\begin{table}[!htb]
\centering
\caption{Endpoint differences at three thresholds in 31 first/last and last/best program-median pairs.}
\label{tab:optimization_sensitivity}
\small
\begin{tabular}{lrr}
\toprule
Threshold & Last faster than first & Last slower than best \\
\midrule
5\% & 19 & 11 \\
10\% & 17 & 8 \\
20\% & 13 & 6 \\
\bottomrule
\end{tabular}
\end{table}

\paragraph{Threshold sensitivity.}
The histories contain 125 distinct programs: 108 observed once, fourteen twice and three three times. \rev{The best program is the fastest of these mostly single timings, so a last/best gap can reflect timing noise as well as a slower program (Appendix~\ref{sec:hpcg_replay}).} Because most programs have a single timing, Table~\ref{tab:optimization_sensitivity} reports the endpoint counts at three thresholds. At 5\%, 10\% and 20\%, there are nineteen, seventeen and thirteen first-to-last gains and eleven, eight and six last/best gaps. Table~\ref{tab:optimization_workflow} also reports application-weighted rates, averaging changes within runs, runs within applications and applications equally, with bootstrap intervals over the seven application means.

\subsection{Candidate Management}\label{sec:trajectory_decisions}
Across 468 trajectories, 19,269 turns contain a modification or restoration/retention label that cites earlier feedback. Modification combines implementation, repair and refinement; restoration/retention combines reverting and retaining or selecting a candidate. Of these turns, 17,537 contain modification only, 1,596 restoration/retention only, and 136 both; multiple qualifying actions in one turn count once. Reading and administrative tools are excluded from modification counts. A turn's feedback source and rationale come only from its qualifying actions.

The cited feedback comes from tests or measurements in 3,979 turns, build execution in 1,699, inspection or profiling in 8,110, other operations in 3,890, and mixed sources in 1,591. When the call that produced the feedback has several facets, test/measurement takes precedence over build execution; reading a timing log counts as inspection. Every feedback link resolves to an earlier observation and the call that produced it.

The narrower restoration/selection label covers 642 tool records in 206 trajectories. Merging simultaneous records gives 583 turns, of which 555 (95.2\%) give a source-bound reason and 542 (93.0\%) cite preceding feedback. These operations include restoring a temporary state, promoting a candidate and making backups. Declared intentions, 303 agent statements and 62 tool records, are counted separately. Among later modification or restoration turns that cite a source-bound public correctness response, 42 in fifteen trajectories cite a passing result and 740 in 159 trajectories cite a failing one.

\subsection{Trajectory Features and Outcomes}\label{sec:trajectory_outcomes}
Four trajectory features were fixed before fitting: presence of local restoration/selection; the share of feedback-linked revision turns containing restoration or retention; presence of a relationship record linking several modification layers; and the share of such records among all relationship groups. Presence features use all 480 trajectories. The restoration share uses the 468 with at least one revision turn, and the layer share the 475 with at least one relationship record. The outcome is per-run reward in percentage points, with all 84 incorrect runs at zero.

Each feature enters a separate regression with a fixed effect for every model--task cell, so its coefficient compares the three runs of one model on one task. Standard errors use a task-cluster CR1 covariance with correction $G/(G-1)\,(N-1)/(N-K)$, where $G=32$ and $K$ is the design rank (161 for the full panel and the layer-share subset, 160 for the restoration-share subset), followed by $t_{31}$ intervals. The feature varies within 100 cells for local restoration/selection, 69 for multi-layer references, and 145 and 150 for the two shares.

All four intervals include zero (Table~\ref{tab:trajectory_outcome_associations}). Before adjustment, the 361 runs with multi-layer references reach $1.25\times$ in 39.6\% and are correct in 80.6\%, versus 26.9\% and 88.2\% for the other 119 runs.

\begin{table}[!htbp]
\centering\small
\caption{\textbf{Process features and final reward.} Each row is a separate within-model--task comparison. Intervals are pointwise 95\% task-cluster CR1 intervals with $t_{31}$ critical values. Variable cells contain variation in the indicated feature.}
\label{tab:trajectory_outcome_associations}
\setlength{\tabcolsep}{4pt}
\begin{tabular*}{\linewidth}{@{\extracolsep{\fill}}lrrr@{}}
\toprule
Feature & Runs & Variable cells & Reward pp [95\% CI] \\
\midrule
Local restoration/selection: present & 480 & 100 & $-1.83$ [$ -6.73, 3.07 $] \\
Restoration/retention share: $+10$ pp & 468 & 145 & $-0.37$ [$ -3.04, 2.31 $] \\
Multi-layer reference: present & 480 & 69 & $+5.62$ [$ -3.44, 14.67 $] \\
Multi-layer group share: $+10$ pp & 475 & 150 & $+0.15$ [$ -2.78, 3.09 $] \\
\bottomrule
\end{tabular*}
\end{table}

\FloatBarrier

\section{Dependencies Across Layers}\label{sec:trajectory_annotations}
Section~\ref{sec:mechanism_controls} estimates how often a change's effect or correctness depends on build settings, runtime settings or the executed code path. The estimate comes from a reviewed random sample (Appendix~\ref{sec:dependency_review}); the annotator's relationship records are used only to locate candidate evidence and to stratify that sample.

\subsection{Relationship Records}\label{sec:trajectory_conditions}
The 23,845 relationship records occur in 475 of the 480 trajectories; the five without records remain in the run denominator. Build requirements occur in 410 trajectories, execution-path dependencies in 440, and at least one of the two in 464. Each record has one of the six types in Table~\ref{tab:trajectory_conditions}, and one trajectory can contain several types. Visible source is the most common basis for build requirements, while execution feedback supports 30.6\% of incompatibility records and 11.5\% of build-requirement records. Performance hypotheses remain agent claims even when their recorded basis is execution feedback, as it is for 153 records; the table reports that basis.

\begin{table}[!htbp]
\centering\small
\caption{\textbf{Optimization relationships across 480 trajectories.} Runs can contain multiple types. Each record has one annotated basis: an agent statement, visible source, or execution feedback.}
\label{tab:trajectory_conditions}
\setlength{\tabcolsep}{4pt}
\begin{tabular*}{\linewidth}{@{\extracolsep{\fill}}lrrrrr@{}}
\toprule
Relationship & Runs & Records & Agent & Source & Feedback \\
\midrule
Build requirement & 410 & 3,956 & 1,284 & 2,216 & 456 \\
Execution-path dependency & 440 & 4,267 & 1,711 & 2,004 & 552 \\
Incompatibility & 440 & 3,923 & 2,326 & 397 & 1,200 \\
Replacement & 451 & 4,331 & 1,790 & 2,224 & 317 \\
Bypass & 432 & 3,819 & 1,832 & 1,371 & 616 \\
Performance hypothesis & 418 & 3,549 & 3,390 & 6 & 153 \\
\bottomrule
\end{tabular*}
\end{table}

For layer summaries, records are grouped by trajectory and cited source turn: the latest cited assistant reasoning or message turn, or otherwise the latest cited turn. This gives 19,191 groups, 705 of which span several annotation chunks. Exact identifiers connect each record's references to annotated actions. A group's layers are the union of the code-editing, build-configuration and runtime-configuration facets it references. At least one layer is linked in 7,737 groups, and 1,814 link two or more (23.4\% of linked groups, 9.5\% of all groups). These multi-layer groups occur in 361 trajectories: 1,245 link code and build, 425 code and runtime, 50 build and runtime settings, and 94 all three.

\subsection{Reviewed Dependency Estimates}\label{sec:dependency_review}
Automatic labels can be accurate case by case and still bias population counts, so we estimate prevalence from a reviewed random sample rather than from annotator counts \citep{swebenchverified,egami2023dsl,angelopoulos2023ppi}. A trajectory is \emph{flagged} if the annotator records a build, execution-path, or bypass dependency supported by execution feedback; 344 of the 480 trajectories are flagged. We sample flagged and unflagged trajectories separately in a fixed pseudorandom order.

A trajectory counts as confirmed if tool output (a build or link error, profile, timing, log, or public-test response) shows that an agent change's effect or correctness depended on a build setting, toolchain, or library; on a runtime setting, input, or thread count; or on which code path the application executed, including a change that is never executed. An agent's own statement, source text alone, or an environment fact unrelated to any change does not count. Each confirmed trajectory is further classified as optimization-related, when the dependency concerns an optimization's speed or correctness, or as an environment workaround, when it concerns only building, loading, or running the program in the evaluation container.

The reviewer searches each sampled trajectory's complete record of tool calls and outputs, not only the annotator's records. Of 40 flagged trajectories, 34 are confirmed (85\%); of 40 unflagged trajectories, 15 are (38\%). Weighting the two strata by their sizes gives an estimated 72\% of all 480 trajectories (95\% interval 63--80\%, normal approximation with finite-population correction). Optimization-related dependencies are confirmed in 24 flagged and 13 unflagged trajectories, giving an estimated 52\% (41--63\%). Flagged trajectories contain about 85\% of the estimated confirmed trajectories. A first-pass review of 40 further flagged trajectories, restricted to at most eight annotator records each, confirms 26; pooling it with the full reviews gives a lower estimate of 64\% (57--71\%). At the record level, 29 of 60 randomly sampled feedback-supported dependency records are confirmed (48\%; build 8/13, execution path 7/21, bypass 14/26). Most rejected records either state the dependency only in the agent's words (11) or describe a comparison between settings or an ordinary bug rather than a dependency (13). The record counts in Appendix~\ref{sec:trajectory_conditions} therefore locate candidate evidence and do not estimate prevalence.

Confirmed cases span the three layers of Section~\ref{sec:mechanism_controls}. They include a thread-count variable that the linear-algebra library ignores, compiler flags that change numerical results (\taskname{ShengBTE}, \taskname{SIESTA}), an algorithm that is slower on small inputs and faster on large ones (\taskname{NUFFT}), an OpenMP change that slows a program at four threads until shared state is made thread-private (\taskname{PRESTO}), an edited \texttt{HPL\_dgemm.c} that leaves the binary byte-identical because a macro maps the call to \texttt{cblas\_dgemm} (\taskname{HPL}), and public tests that return before the edited code runs.

\section{Controlled Experiments}\label{sec:controlled_replay}\label{sec:completed_controls}
Each experiment replays fixed programs under the unchanged private verifier and varies one set of factors. \taskname{GROMACS} varies build and runtime settings on identical source, \taskname{QiboTN} switches three code changes on and off in a full factorial design, and \taskname{HPCG-derived} compares each run's fastest public-test candidate with its final submission. All replays run on dedicated physical cores within one NUMA node, with SMT siblings excluded; background services remain active.

\subsection{\taskname{GROMACS}: One Source, Four Configurations}\label{sec:gromacs_control}
The task rebuilds the submitted molecular-simulation source according to \texttt{workflow.json} and times the complete prescribed \texttt{mdrun}. In the replayed run, the final submission differs from the pristine source only in \texttt{workflow.json}, so the whole gain lies in build and runtime configuration. We replay the identical source under four combinations of SIMD and thread settings on four physical cores. The one-thread variants keep this allocation, and both thread counts reuse the same clean build for each SIMD setting.

Speedups of the same source range from 0.289$\times$ to 2.123$\times$ across the four configurations, all of which pass the numerical and replay checks on six private physical profiles with three interleaved timing repetitions per profile (Table~\ref{tab:gromacs_control}). With SIMD disabled and four threads, the unchanged source measures 0.992$\times$. AVX2 with four threads reaches 2.123$\times$; with one thread, AVX2 gives 0.709$\times$ and disabled SIMD 0.289$\times$.

\begin{table}[!ht]
\caption{\textbf{The same \taskname{GROMACS} source ranges from 0.289$\times$ to 2.123$\times$.} Runtime is the sum of six workload medians. Each row has its own interleaved measurement of the fixed SIMD-disabled, four-thread reference. All rows pass correctness.}
\label{tab:gromacs_control}
\centering\small
\begin{tabular}{@{}lrrrr@{}}
\toprule SIMD & Threads & Reference (s) & Candidate (s) & Speedup\\\midrule
Disabled & 1 & 30.58 & 105.70 & 0.289$\times$\\
Disabled & 4 & 30.48 & 30.72 & 0.992$\times$\\
AVX2 & 1 & 30.69 & 43.31 & 0.709$\times$\\
AVX2 & 4 & 30.80 & 14.51 & 2.123$\times$\\
\bottomrule
\end{tabular}
\end{table}

\subsection{\taskname{QiboTN}: Interacting Code Changes}\label{sec:qibo_control}
The controlled program is DeepSeek-V4-Flash's final \taskname{QiboTN} submission \texttt{5KUtPHw}. We vary broad dispatch to a fast exact simulator (B), reuse of a previously computed state for an observable (R), and deferred loading of tensor libraries (L). Level one keeps the submission's behavior and zero disables only that feature. All interventions edit the same backend file and leave the other 66 files unchanged, so the all-off variant still contains the submission's other changes. Each of the eight variants receives three complete application evaluations in a fixed random order.

\paragraph{Workload and checks.}
Each round runs six fresh-process variational circuits with 9--14 qubits, with identical inputs across variants within a round and different fixed seeds across rounds. We check the full complex state against an independent calculation, its norm, all single-qubit expectations, a symbolic observable, selected probabilities and exact numerical replay. Every recipe runs each role twice, in reference--candidate and candidate--reference order. The reported time sums six per-recipe median wall times, including process startup, imports, execution and full output. The environment uses eight physical cores, a 32\,GiB limit, a fixed container image and no network; the source and tests are checked for changes before and after each evaluation. All 24 evaluations pass, giving 576 process timings.

\begin{table}[!ht]
\caption{\textbf{Deferred loading (L) helps \taskname{QiboTN} only when broad dispatch (B) is on.} Each row is one variant of the eight-variant factorial design; 1 keeps the submission's behavior and 0 disables it. Candidate time sums six per-recipe median wall times. Gain is the all-off candidate time divided by the variant's candidate time within the same round; share is the fraction of the all-on runtime saving that the variant recovers. Ranges span three rounds with different circuit seeds. All 24 evaluations pass correctness.}
\label{tab:qibo_control}
\centering\small
\begin{tabular}{@{}cccrrr@{}}
\toprule B & R & L & Candidate (s) & Gain over all-off & Share of saving (\%)\\\midrule
0 & 0 & 0 & 76.99--77.12 & 1.000$\times$ & 0.00\\
1 & 0 & 0 & 50.86--51.00 & 1.512--1.515$\times$ & 37.76--37.93\\
0 & 1 & 0 & 76.94--77.21 & 0.999--1.002$\times$ & $-0.14$ to $0.18$\\
0 & 0 & 1 & 75.96--76.02 & 1.014--1.015$\times$ & 1.50--1.60\\
1 & 1 & 0 & 50.89--50.93 & 1.512--1.514$\times$ & 37.79--37.88\\
1 & 0 & 1 & 7.94--8.04 & 9.592--9.711$\times$ & 99.87--100.08\\
0 & 1 & 1 & 75.98--76.01 & 1.013--1.015$\times$ & 1.44--1.61\\
1 & 1 & 1 & 7.95--8.01 & 9.614--9.699$\times$ & 100.00\\
\bottomrule
\end{tabular}
\end{table}

\paragraph{Results.}
Deferred loading (L) gives a large gain only when broad dispatch (B) is on. With R off, adding L to the B-on variant improves runtime by 6.343--6.409$\times$, whereas adding L to the all-off variant improves it by 1.014--1.015$\times$ (Table~\ref{tab:qibo_control}). Averaged over both R levels on the log-runtime scale, the B--L interaction is a ratio of gains of 6.278--6.307. All three features together accelerate the all-off variant by 9.614--9.699$\times$, and no single feature meets the predeclared criterion of recovering 90\% of this combined saving in every round. Removing B from the full submission slows it by 9.490--9.560$\times$, and removing L by 6.359--6.405$\times$, so the knockout effects overlap rather than add. The source explains the dependency: deferred loading avoids importing the tensor library only when the execution path does not need it. When B is off, these recipes never reach the state-reuse path, which makes R an unreachable-path control.

Because the rounds use different circuit seeds, the ranges include both input and timing variation. Median within-recipe differences between candidate pairs range from 0.31\% to 1.33\% across variants, and baseline aggregate times range from 75.793 to 76.192 seconds. The baseline is the supplied pristine \taskname{QiboTN} source with its original application runner, whereas the all-off variant keeps the submission's other edits.

\subsection{\taskname{HPCG-derived}: Selected Candidates Versus Final Submissions}\label{sec:hpcg_replay}
For seven \taskname{HPCG-derived} runs, we replay the candidate with the best public-test runtime in each run alongside its final submission. The replay set contains 18 candidate instances covering 16 distinct programs: the seven final submissions, the six best public-test candidates that differ from them, and five further candidates selected from the public feedback available at earlier budget cutoffs. Three outer rounds produce 54 complete private evaluations, 324 case checks, and 972 baseline and 972 candidate process timings; every evaluation passes. All seven final submissions solve the task's constant-coefficient stencil with a sine-transform direct solver for both right-hand sides.

The same algorithm family still yields different runtimes. In six runs, the candidate with the best public-test runtime differs from the final submission: two final submissions are faster in every round, two are slower in every round, and two change direction across rounds. In the seventh run the two are identical. A single timing comparison can therefore misstate which of two candidates is faster.

Replay conditions also shift absolute speedups. The replay fixes a new right-hand-side seed and runs on a different allocation from the original runs. Under these conditions, one run's recorded final speedup of 63.36$\times$ becomes 19.73$\times$, mostly through a change in candidate time. Capping that program at four threads gives a three-run median of 20.28$\times$ with overlapping timing ranges, so thread oversubscription alone does not explain the difference. The replay does not change official scores.

\section{Case Records}\label{sec:mechanism_evidence}
The cases below come from the 480 analyzed runs and are grouped by the observation in Section~\ref{sec:analysis} they illustrate. We treat agent statements as hypotheses, and tool output and submitted source as observations. Step numbers refer to the original trajectory.

\subsection{Repairing a Candidate After a Failed Check}\label{sec:feedback_decisions}
\paragraph{\taskname{HPCG-derived}: an indexing error found by residuals.}
The task solves a fixed three-dimensional 27-point Dirichlet system for two right-hand sides; correctness requires small solution error, small independently computed residuals and reproducible vectors. The operator admits transform-based direct solvers. In Hy3 run \texttt{u2x9KjE}, the agent's transform-based candidate produces relative residuals of about 1.6--1.9. After the agent changes the slice stride in \texttt{apply\_y} to \texttt{ny*nx} and rebuilds, the same checker reports residuals of $3.0\times10^{-15}$ to $9.1\times10^{-15}$.

\paragraph{\taskname{QuEST}: an operation applied twice.}
The task simulates state-vector and density-matrix quantum circuits and checks complete states and probabilities. In Hy3 run \texttt{r2HQQna}, step 146 reports a maximum amplitude discrepancy of 0.644 for the agent's fused density-matrix path. At steps 174--175 the agent attributes the discrepancy to applying the ket operation twice and changes the dispatch guard. The patch introduces a build error that the agent repairs by step 180. After further fixes to its test harness, step 201 reports no differing local cases, and step 208 returns a public correctness pass.

\subsection{Keeping a Slower Candidate After Numerical Checks}\label{sec:su2_selection}
\taskname{SU2} runs 250 iterations of a transonic airfoil simulation and must write five complete flow-field snapshots within a field-error tolerance of $3\times10^{-3}$. In DeepSeek-V4.1-Flash run \texttt{evDkccR}, the agent compares two preconditioner settings. At step 849, Jacobi-12 gives worst relative field $L_2$ errors of $2.096\times10^{-4}$ and $1.759\times10^{-4}$ on the two public profiles, and Jacobi-15 raises the first to $1.489\times10^{-3}$ while lowering the second. At step 850, four interleaved local timings per setting give medians of 3.030\,s for Jacobi-12 and 4.297\,s for LU-SGS with five inner iterations.

At step 851 the agent rejects Jacobi, citing errors that change non-monotonically with the iteration count and uncertainty about other inputs. It keeps LU-SGS, checks that its configuration differs from the reference only in the preconditioner thread setting, and obtains two public correctness passes at step 852. The final submission scores reward 0.456. The discarded Jacobi candidate was not replayed privately, so the private-verifier speed given up by this choice remains unmeasured.

\subsection{Changes the Application Does Not Execute}\label{sec:execution_path_cases}
\paragraph{\taskname{HPL}: an optimized kernel bypassed by the build.}
The task runs a complete MPI dense linear solve, from input distribution to solution export and checks. In DeepSeek-V4-Flash run \texttt{hiruimA}, the agent optimizes \texttt{HPL\_dgemm.c}. A \texttt{cmp} check reports the stock and optimized executables as identical, and the expected kernel symbol is absent; a subsequent symbol inspection shows an unresolved \texttt{cblas\_dgemm} call. The agent then moves its optimization into a strong CBLAS override with a fallback. A source listing read late in the trajectory still contains the override, and the agent's closing summary lists it among the retained changes. Because the final submission also changes application I/O, its reward of 0.290 does not isolate the override's speedup; the case documents the execution-path check.

\paragraph{\taskname{QiboTN}: a local test that misses the changed route.}
The task executes quantum circuits and must return complete states and observables. In Grok 4.6 run \texttt{fXyyQog}, the public workflow passes before the agent changes the tensor-network fallback. The next local smoke run uses the dense-state route and therefore does not exercise the changed fallback. The final submission is correct, so the case concerns what the executed test covered and does not indicate a faulty fallback.

\subsection{Build Settings That Change Numerical Output}\label{sec:build_numerics_case}
\taskname{GPAW} converges an electronic state, takes two geometry steps, writes a wavefunction checkpoint and resumes in a fresh process. In DeepSeek-V4-Flash run \texttt{Ykcn7Dy}, an x86-64-v3 build produces different output and checkpoint files. The agent first attributes the difference to vectorization and then builds with fused multiply--add disabled. For that build, the same file comparison that had reported differences now reports identical initial, geometry and checkpoint files, and identical resumed output. The agent revises its explanation, calls the measured speed gain marginal, and reports keeping its earlier build. Here diagnosis and selection diverge: the agent reports that the repaired change was not worth keeping, and the source tree does not independently confirm that choice.

\clearpage
\section{Optimization Trajectories and Final Outcomes}\label{sec:trajectory_atlas}
\clearpage

\end{document}